\documentclass[10pt,journal,cspaper,compsoc]{IEEEtran}
\ifCLASSOPTIONcompsoc
\else
\fi
\ifCLASSINFOpdf
\else
\fi
\usepackage{graphicx} 
\usepackage{subfigure}

\usepackage{lettrine}

\usepackage{algorithm}
\usepackage{algorithmic}

\usepackage{amssymb}
\usepackage{amsthm}
\usepackage{amsmath}
\usepackage{amsbsy}
\usepackage[sort&compress,square,numbers]{natbib}

\newtheorem{theorem}{Theorem}
\newtheorem*{theorem*}{Theorem}
\newtheorem{corollary}{Corollary}
\newtheorem*{corollary*}{Corollary}
\newtheorem{definition}{Definition}
\newtheorem{lemma}{Lemma}
\newtheorem*{lemma*}{Lemma}
\newtheorem{proposition}{Proposition}
\newtheorem*{proposition*}{Proposition}

\allowdisplaybreaks

\begin{document}
%
\title{Multi-view Intact Space Learning}
%
%
%
%

\author{Chang Xu,
        Dacheng Tao,~\IEEEmembership{Fellow,~IEEE,}
        Chao Xu
\IEEEcompsocitemizethanks{
\IEEEcompsocthanksitem Chang Xu and Chao Xu are with the Key Laboratory of Machine
Perception (Ministry of Education), School of Electronics Engineering
and Computer Science, Peking University, Beijing 100871, China (e-mail:
changxu1989@gmail.com; xuchao@cis.pku.edu.cn).\protect\\
\IEEEcompsocthanksitem
D. Tao is with the Centre for Quantum Computation \& Intelligent Systems and the Faculty of Engineering \& Information Technology, University of Technology, Sydney, 235 Jones Street, Ultimo, NSW 2007, Australia (email: dacheng.tao@uts.edu.au; Office number: +61 2 9514 1829 ; Fax number: +61 2 9514 4517).

}
\thanks{\textcopyright 20XX IEEE. Personal use of this material is permitted. Permission from IEEE must be obtained for all other uses, in any current or future media, including
reprinting/republishing this material for advertising or promotional purposes, creating new collective works, for resale or redistribution to servers or lists, or reuse of any copyrighted component of this work in other works.}}

%
%

\markboth{IEEE TRANSACTIONS ON PATTERN ANALYSIS AND MACHINE INTELLIGENCE, VOL. X, NO. X, MONTH 20XX}%
{Shell \MakeLowercase{\textit{\textit{et al.}}}: Bare Demo of IEEEtran.cls for Computer Society Journals}
%


\IEEEcompsoctitleabstractindextext{%
\begin{abstract}
It is practical to  assume that an individual view is unlikely to be sufficient for effective multi-view learning. Therefore, integration of multi-view information is both valuable and necessary. In this paper, we propose the Multi-view Intact Space Learning (MISL) algorithm, which integrates the encoded complementary information in multiple views to discover a latent intact representation of the data. Even though each view on its own is insufficient, we show theoretically that by combing multiple views we can obtain abundant information for latent intact space learning.
Employing the Cauchy loss (a technique used in statistical learning) as the error measurement strengthens robustness to outliers. We propose a new definition of multi-view stability and then derive the generalization error bound based on multi-view stability and Rademacher complexity, and show that the  complementarity between multiple views is beneficial for the stability and generalization. MISL is efficiently optimized using a novel Iteratively Reweight Residuals (IRR) technique, whose convergence is theoretically guaranteed. Experiments on synthetic data and real-world datasets demonstrate that MISL is an effective and promising algorithm for practical applications.
\end{abstract}

}

\maketitle

\IEEEdisplaynotcompsoctitleabstractindextext

%
\IEEEpeerreviewmaketitle

\section{Introduction}

\lettrine{M} ost of the data used in video surveillance, social computing, and environmental sciences are collected from diverse domains or obtained from various feature extractors. These data are heterogeneous, because their variables can be naturally partitioned into groups. Each variable group is referred to as a particular view, and the multiple views for a particular problem can take different forms. For example, a sparse camera network containing multiple cameras is used for person re-identification and understanding global activity through color descriptors, local binary patterns, local shape descriptors, slow features and spatial temporal contexts.

The information obtained from an individual view cannot comprehensively describe all examples.  It has therefore become popular to leverage the information derived from the connections and differences between multiple views to better describe the objects, which has resulted in multi-view learning algorithms that integrate multiple features from diverse views (or simply multi-view features).

Recently, numbers of multi-view learning algorithms have been designed and successfully applied to various computer vision and intelligent system problems \cite{yuan2012multi, xiang2013multi}. Co-training \cite{blum1998combining} is one of the earliest semi-supervised schemes for multi-view learning. It trains alternately to maximize the mutual agreement on two distinct views of the unlabeled data. Many variants \cite{nigam2000analyzing, muslea2002active, Co-regularization, kumar2011co, muslea2006active, multi_view_clustering, kumar2010co, yu2011bayesian}, such as co-EM \cite{nigam2000analyzing} and co-regularization \cite{Co-regularization, kumar2011co}, have since been developed. Their success is relied on the assumption  that the two sufficient and redundant views are conditional independent to the other given the class label. However, this assumption tends to be too rigorous for many practical applications, and thus some alternative  assumptions have been studied. \cite{abney2002bootstrapping}  showed that weak dependence can also guarantee successful co-training. \cite{balcan2004co} proved a weaker assumption called $\epsilon$-expansion was sufficient for iterative co-training to succeed. After that, 
Wang and Zhou conducted a series of in-depth analyses and revealed some interesting properties of co-training, including the large-diversity of classifiers \cite{wang2007analyzing, zhou2005tri}, label propagation over two views \cite{Co-traning-label} and co-training with insufficient views \cite{wang2013co}.


Multiple kernel learning (MKL) was originally developed to control the search space capacity of possible kernel matrices to achieve good generalization but has been widely applied to problems involving multi-view data \cite{ji2008multi, tang2009multiple}. This is because kernels in MKL naturally correspond to different views and combining kernels either linearly or non-linearly improves learning performance, especially when the views are assumed to be independent. \cite{lanckriet2002learning, lanckriet2004learning} formulated MKL as a semi-definite programming problem. \cite{bach2004multiple}  treated MKL as a second order cone program problem and developed an SMO algorithm to efficiently obtain the optimal solution. \cite{sonnenburg2006general, sonnenburg2006large} developed an efficient semi-infinite linear program and made MKL applicable to large scale problems. \cite{rakotomamonjy2007more, rakotomamonjy2008simplemkl} proposed simple MKL by exploring an adaptive 2-norm regularization formulation. \cite{szafranski2010composite, xu2010simple} constructed the connection between MKL and group-LASSO to model group structure.

A number of works exploit the shared latent subspace across diverse views, such as canonical correlation analysis (CCA) \cite{chaudhuri2009multi, kakade2007multi, sun2011canonical}, its kernel  extension \cite{akaho2006kernel}, its probabilistic interpretation \cite{bach2005probabilistic}, and its sparse formulation \cite{archambeau2009sparse}. Recently, other methodologies have been proposed for this task: \cite{lawrence2004gaussian, shon2005learning} used Gaussian process to discover latent variable model shared by multi-view data; \cite{memisevic2006kernel, memisevic2012shared} found the joint embedding for multi-view data by maximizing mutual information; these techniques are particularly effective for modeling the correlations between different views. To simultaneously account for the dependencies and independencies of different input views, various methods have been introduced that factorize the latent space into a shared part common to all views and a private part for each view \cite{salzmann2010factorized, jia2010factorized}. By considering the side information, the recent work of max-margin Harmonium (MMH) \cite{chen2012large} showed that applying the large-margin principle to learn subspace shared by multi-view data is more suitable for prediction.

However, existing multi-view learning methods have their own limitations. First, since an individual view  is insufficient for learning, integration of multi-view information is necessary and valuable; however, besides several works concentrating on co-training style algorithms \cite{ando2007two,wang2013co}, the issue of single-view insufficiency has not been clearly addressed and comprehensively studied. Second, the term``intact'' means \emph{complete} and \emph{not damaged} in Merriam-Webster, which are exactly the two favorable properties  we wish to possess in the latent intact space. However, most of the existing multi-view learning algorithms fail to discover latent intact spaces, due to the information loss in learning from insufficient views or the influence of noises in insufficient views.
Finally, there is a demand on the theoretical supports to guarantee the performance of multi-view learning.

In this paper, we assume that while each individual view only captures partial information, all the views together possess redundant information of the object. In contrast to most existing multi-view learning models that assume view sufficiency, we propose a Multi-view Intact Space Learning (MISL) algorithm to address insufficiency in each individual view and to integrate the encoded complementary information. The new view functions in the MISL algorithm are rigorously studied.
To enhance the robustness of the model, we measure the reconstruction error from different views using the Cauchy loss, which is robust to outliers and has an optimal breakdown point compared with conventional $L_{2}$ and $L_{1}$ losses \cite{he2000breakdown}. To solve the two sub-problems w.r.t. the view generation functions and the intact space derived from the optimization problem, we develop an Iteratively Reweight Residuals (IRR) optimization technique, which is efficiently implemented and has guaranteed convergence. Although each view only captures partial information of the latent intact space, MISL theoretically guarantees that given enough views the latent intact space can be approximately restored.
We introduce a new definition of ``multi-view stability'' to analyze the robustness of the proposed algorithm. Moreover, we derive the generalization error bound based on the multi-view stability and Rademacher complexity, and show that the complementarity of multiple views can improve  the multi-view stability and  the generalization. Finally, we conduct experiments to explicitly illustrate the view insufficiency assumption and  the robustness of the proposed algorithm and show, using real-world datasets, that our approach can accurately discover an intact space for the subsequent classification tasks.

The rest of the paper is organized as follows. In Section 2, we formulate the multi-view learning problem and propose the MISL algorithm. The optimization method is presented in Section 3 and theoretical analysis is given in Section 4. Section 5 presents the  experimental results, and Section 6 concludes the paper. The detailed proofs of the theoretical results are in Section 7.

\section{Problem Formulation}

\begin{figure}[tb]
\begin{center}
   \includegraphics[width=\columnwidth]{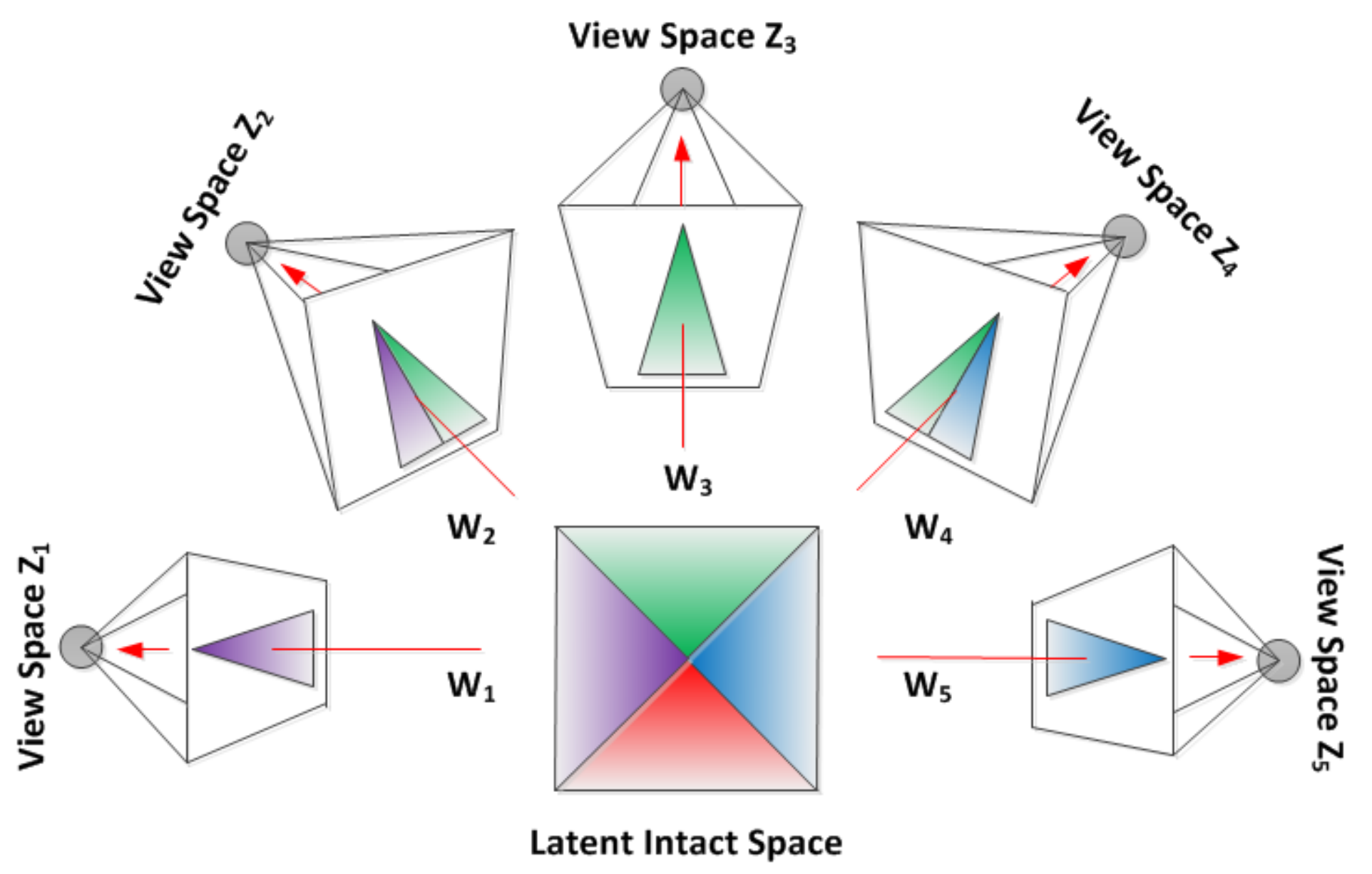}
\end{center}\vskip -0.1in
   \caption{View insufficiency assumption.}\vskip -0.2in
\label{fig:scn}
\end{figure}

View sufficiency is usually not guaranteed in practice. By contrast, we assume ``view insufficiency'' that each view only captures partial information but all the views together carry redundant information about the latent intact representation (shown in Figure \ref{fig:scn}). Many practical problems support this assumption. For example, in a camera network, cameras are placed in public areas to predict potentially dangerous situations in time to take necessary action. However, each camera alone captures insufficient information and thus cannot comprehensively describe the environment, which can only be fully recovered by integrating multiple data from all the cameras.

In multi-view learning, an example $x$ is represented by multi-view features $z^{v}$, where $m$ is the number of views and $1\leq v\leq m$. Supposing $x$ is the latent intact representation, each view $z^{v}$ is a particular reflection of the example, and obtained from the view generation function $f^{v}$ on $x$,
\begin{equation}\label{eq:noise}
    z^{v} = f^{v}(x) + \epsilon^{v},
\end{equation}
where $\epsilon^{v}$ is the view-dependent noise. According to the view insufficiency assumption, we know that the function $f^{v}(x)$ is non-invertible, so we cannot recover $x$ from $z^{v}$ even given the view function $f^{v}: \mathcal{X}\rightarrow \mathcal{Z}^{v}$. For a linear function $f^{v}(x)=W_{v}x$, non-invertibility implies that $W_{v}$ is not column full-rank.

Hence, our objective is to learn a series of view generation functions $\{W_{v}\}_{v=1}^{m}$ to generate multi-view data points from a latent intact space $\mathcal{X}$. A straightforward approach is minimizing the empirical risk over  $\{z^{v}-f^{v}(x)\}_{v=1}^{m}$ using the $L_{1}$ or $L_{2}$ loss. Given Eq. (\ref{eq:noise}), the noise in different views  seriously influences the discovery of the optimal view generation functions and the latent intact space. However, as thoroughly studied in robust statistics \cite{rey1983introduction},  neither $L_{1}$ nor $L_{2}$ loss is robust to outliers, and thus the  performance of multi-view learning will be seriously degraded.



\subsection{Robust Estimators}

M-estimator is popular in robust statistics. Let $r_{i}$ be the residual of the $i$-th data point, i.e., the difference between the $i$-th observation and its fitted values. The standard least-squares method tries to minimize $\sum_{i}r_{i}^{2}$, which is unstable if outliers are present, and which has a strong effect to distort the estimated parameters. M-estimators try to reduce the effect of outliers by replacing the squared residuals $r_{i}^{2}$ with another function of residuals
\begin{equation}
    \min \; \sum_{i}\rho(r_{i}),
\end{equation}
where $\rho$ is a symmetric, positive-definite function with a unique minimum at zero and chosen to be less increasing than the square function. The corresponding influence function is defined as
\begin{equation}\label{}
    \psi(x) = \frac{\partial \rho(x)}{x},
\end{equation}
which measures the influence of a random data point on the value of the parameter estimate.

\begin{figure}[tb]
\begin{center}
   \includegraphics[width=\columnwidth]{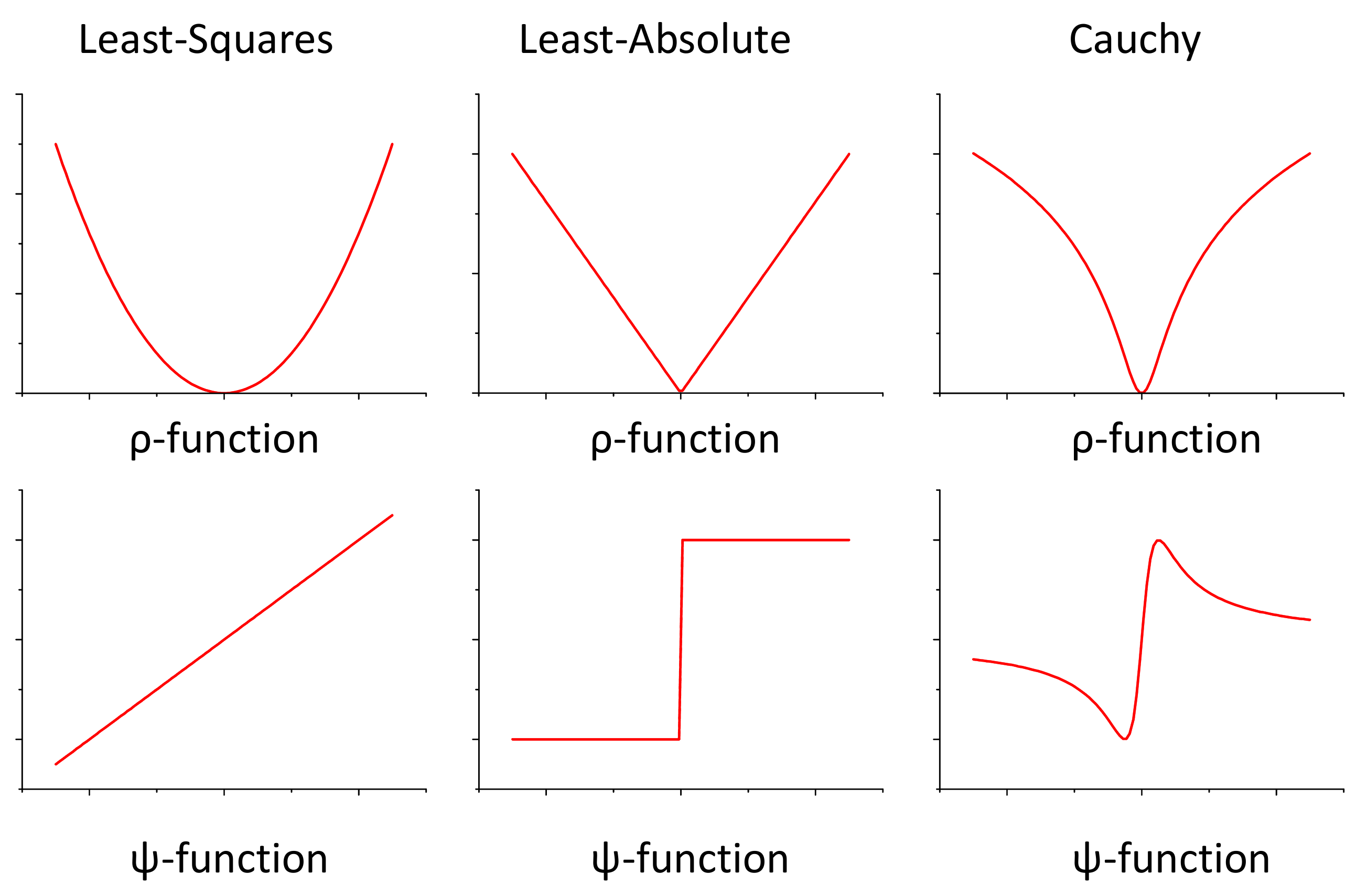}
\end{center} \vskip -0.1in
   \caption{Example robust estimators.}\vskip -0.2in
\label{fig:estimator}
\end{figure}

As shown in Figure \ref{fig:estimator}, for the $L_{2}$ estimator (least-squares) with $\rho(x)=x^{2}/{2}$, the influence function is $\psi(x)=x$; that is, the influence of a data point on the estimate increases linearly with the size of its error. This confirms the non-robustness of the least-squares estimate. Although the $L_{1}$ (absolute value) estimator with $\rho(x)=|x|$ reduces the influence of large errors, its influence function has no cut-off. When an estimator is robust, it is inferred that the influence of any single observation is insufficient to yield a significant offset. Cauchy estimator  has been shown to own this valuable property
\begin{equation}\label{}
    \rho(x) = \log(1+(x/c)^{2})
\end{equation}
along with the upper bounded influence function
\begin{equation}\label{}
    \psi(x) = \frac{2x}{c^{2}+x^{2}}.
\end{equation}

Furthermore,  Cauchy estimator \cite{mizera2002breakdown} theoretically has a breakdown point of nearly $50\%$, which means that nearly half of the observations (e.g., arbitrarily large observations) can be incorrect before the estimator gives an incorrect (e.g., arbitrarily large) result. Therefore, we deploy the Cauchy estimator in MISL.

\subsection{Multi-view Intact Space Learning}

We consider a  multi-view training sample $\mathcal{D}=\{z_{i}^{v}| 1\leq i \leq n, \; 1 \leq v \leq m \}$ whose view number is $m$ and sample size is $n$. The reconstruction error over the latent intact space $\mathcal{X}$ can be measured using the Cauchy loss
\begin{equation}\label{eq:recon}
    \frac{1}{mn}\sum_{v=1}^{m}\sum_{i=1}^{n}\log{\left(1+\frac{\|z_{i}^{v}-W_{v}x_{i}\|^{2}}{c^{2}}\right)},
\end{equation}
where $x_{i}\in\mathbb{R}^{d}$ is a data point in the latent intact space $\mathcal{X}$, $W_{v}\in\mathbb{R}^{D_{v}\times d}$ is the $v$-th view generation matrix, and $c$ is a constant scale parameter.

Moreover, we adopt some regularization terms to penalize the latent data point $x$ and the view generation matrix $W$. Finally, the resulting objective function can be written as
\begin{equation}\label{eq:p}
\begin{split}
  \min_{x, W} \; & \frac{1}{mn} \sum_{v=1}^{m}\sum_{i=1}^{n}\log{\left(1+\frac{\|z_{i}^{v}-W_{v}x_{i}\|^{2}}{c^{2}}\right)} \\
    & +C_{1}\sum_{v=1}^{m}\|W_{v}\|_{F}^{2}+C_{2}\sum_{i=1}^{n}\|x_{i}\|_{2}^{2}\\
\end{split}
\end{equation}
where $C_{1}$ and $C_{2}$ are non-negative constants that can be determined using cross validation. Problem (\ref{eq:p}) jointly models the relationships between the latent intact space $\mathcal{X}$ and each view space $\mathcal{Z}^{v}$ using a robust approach. It can be expected that by solving this problem with an input of multiple insufficiency views, a series of view generation functions and a latent intact space can be found that represents the object in its entirety.

At inference, given a new multi-view example $\{z^{1}, \cdots, z^{m}\}$, the corresponding data point $x$ in the intact space can be obtained by solving the problem
\begin{equation}
    \min_{x}\; \frac{1}{m} \sum_{v=1}^{m}\log{\left(1+\frac{\|z^{v}-W_{v}^{*}x\|^{2}}{c^{2}}\right)}+C_{2}\|x\|_{2}^{2},
\end{equation}
where $W^{*}_{v}$ is the optimal view generation function.

\subsection{Kernel Extension}

When the view space $Z$ lies in an infinite-dimensional Hilbert space, there exists a nonlinear mapping $\phi: \mathbb{R}^{d}\rightarrow \mathcal{H}$, such that $k(z,z_{i})=\langle\phi(z),\phi(z_{i})\rangle$. Furthermore, we denote the learned projection function $W$ in the feature space as $\phi(W)$. Assuming atoms of $W$ lie in the space spanned by the input data, we can write $\phi(W)=\phi(Z)A$, where $A$ is the atom representation matrix and $\phi(Z)=[\phi(z_{1}), \cdots, \phi(z_{n})]$. The kernel extension of Problem (\ref{eq:p}) can then be obtained through
{\small
\begin{equation}\nonumber
\begin{split}
  &\|z_{i}^{v}-W_{v}x_{i}\|_{\mathcal{H}}^{2}= \langle\phi(z_{i}^{v})-\phi(W_{v})x_{i}, \phi(z_{i}^{v})-\phi(W_{v})x_{i}\rangle \\
    &= \langle\phi(z_{i}^{v}),\phi(z_{i}^{v})\rangle-2\langle\phi(z_{i}^{v}),\phi(W_{v})x_{i}\rangle+\langle\phi(W_{v})x_{i},\phi(W_{v})x_{i}\rangle\\
    &=k(z_{i}^{v},z_{i}^{v})-2\sum_{j=1}^{n}k(z_{i},z_{j})A_{v}x_{i}+x_{i}^{T}A_{v}^{T}K_{Z^{v}}A_{v}x_{i},
\end{split}
\end{equation}
}\noindent
and
\begin{equation}\label{}
   \begin{split}
    \|W_{v}\|^{2}_{\mathcal{H}} =&  \langle\phi(Z^{v})A_{v},\phi(Z^{v})A_{v}\rangle=A_{v}^{T}K_{Z^{v}}A_{v}
    \end{split}
\end{equation}
where $K_{Z^{v}}$ and $A_{v}$ are the kernel matrix and the atom matrix of view-$v$ respectively.
The kernelized problem can be optimized using the same technique as the linear MISL defined by Eq. (7).


\section{Optimization}

Problem (\ref{eq:p}) can be decomposed into two sub-problems over the view generation function $W$ and the latent intact space $\mathcal{X}$  using the alternating optimization method. Inspired by the generalized Weiszfeld's method \cite{voss1980linear}, we develop an Iteratively Reweight Residuals (IRR) algorithm to efficiently optimize these two subproblems.
%

Given fixed view generation functions $\{W_{v}\}_{v=1}^{m}$, Eq. (\ref{eq:p}) can be minimized over each latent point $x$ in the latent intact space $\mathcal{X}$,
\begin{equation}\label{eq:j1}
    \min_{x}\; \mathcal{J}=\frac{1}{m}\sum_{v=1}^{m}\log{\left(1+\frac{\|z^{v}-W_{v}x\|^{2}}{c^{2}}\right)}+C_{2}\|x\|_{2}^{2}.
\end{equation}

Setting the gradient of $\mathcal{J}$ with respect to $x$ to $0$, we have

\begin{equation}\label{eq:d}
 \sum_{v=1}^{m}-\frac{2W_{v}^{T}(z^{v}-W_{v}x)}{c^{2}+\|z^{v}-W_{v}x\|_{2}^{2}}+2mC_{2}x=0,
\end{equation}
which can be rewritten as
{\scriptsize
\begin{equation}\label{eq:d2}
      \left(\sum_{v=1}^{m}\frac{W_{v}^{T}W_{v}}{c^{2}+\|z^{v}-W_{v}x\|_{2}^{2}}+mC_{2}\right)x = \sum_{v=1}^{m}\frac{W_{v}^{T}z^{v}}{c^{2}+\|z^{v}-W_{v}x\|_{2}^{2}},
\end{equation}
}\noindent
where $r^{v}=z^{v}-W_{v}x$ is referred to as the  residual of the example $x$ on each view.
A weight function is then defined as
\begin{equation}\label{eq:w}
    Q = \left[\frac{1}{c^{2}+\|r^{1}\|^{2}},\cdots, \frac{1}{c^{2}+\|r^{m}\|^{2}}\right],
\end{equation}
which can be used to reduce the influence of outliers and adjust the errors introduced by different views.
Based on Eqs. (\ref{eq:d2}) and (\ref{eq:w}), we have
\begin{equation}\label{eq:update}
    x = \left(\sum_{v=1}^{m}W_{v}^{T}Q_{v}W_{v}+mC_{2}\right)^{-1}\sum_{v=1}^{m}W_{v}^{T}Q_{v}z^{v}.
\end{equation}

Considering $Q$ depends on $x$, we thus iteratively update $x$ using Eq. (\ref{eq:update})  with an initial estimate until convergence. The iterative procedure is described in algorithm \ref{alg:irr}.
\begin{algorithm}[thb]
   \caption{Iteratively Reweighted Residuals (IRR)}
   \label{alg:irr}
\begin{algorithmic}
   \STATE {\bfseries Input:} $\{z_{1},\cdots, z_{m}\}$, $\{W_{v}\}_{v=1}^{m}$, and $x^{0}$
   \STATE Initial residuals $\{r_{v}\}_{v=1}^{m}$ are found using $x^{0}$
   \FOR{$k=1,\cdots$}
   \STATE Weight function $Q$ is chosen through Eq. (\ref{eq:w})
   \STATE Using Eq. (\ref{eq:update}) to obtain the estimate $x^{k}$
   \STATE Update the residuals $\{r_{v}\}_{v=1}^{m}$
   \IF{the estimates of $x$ converge}
   \STATE break
   \ENDIF
   \ENDFOR
   \STATE {\bfseries Output:} $x^{k}$
\end{algorithmic}
\end{algorithm}

By fixing  all data points in the intact space $\mathcal{X}$, Eq. (\ref{eq:p}) is reduced to the minimization over each view generation function $W$
\begin{equation}\label{eq:j2}
     \min_{W}\; \mathcal{J}=\frac{1}{n}\sum_{i=1}^{n}\log{\left(1+\frac{\|z^{i}-Wx_{i}\|^{2}}{c^{2}}\right)}+C_{1}\|W\|^{2}.
\end{equation}

Given the residual $r_{i}=z^{i}-Wx_{i}$ and the weight function on the training data
\begin{equation}\label{eq:w2}
    Q = \left[\frac{1}{c^{2}+\|r_{1}\|^{2}},\cdots, \frac{1}{c^{2}+\|r_{n}\|^{2}}\right],
\end{equation}
we can update the projection function $W$ by
\begin{equation}\label{eq:update_2}
    W = \sum_{i=1}^{n}z_{i}Q_{i}x_{i}^{T}
    \left(\sum_{i=1}^{n}x_{i}Q_{i}x_{i}^{T}+nC_{1}\right)^{-1}.
\end{equation}

Similar to the optimization over the latent comprehensive space $\mathcal{X}$, we can also estimate $W$ by Algorithm \ref{alg:irr}.

\section{Theoretical Analysis}

In this section, we analyze the convergence of the optimization technique, present a new definition of ``multi-view stability'', and then derive the generalization error bound of the proposed multi-view learning algorithm.  The detailed proofs are given in Section 7.

\subsection{Convergence Analysis}

We employ the majorize-minimize framework from \cite{voss1980linear} to analyze the convergence of the IRR algorithm. The key idea of this framework is to globally approximate $\mathcal{J}$ using a sequence of quadratic functions. Taking the subproblem over $x$ as an example, after having found $x^{k}$, we can construct a quadratic function $\psi(x;x^{k})$ to upper bound $\mathcal{J}(x)$ such that the following conditions hold:
\begin{equation}\label{}
    \begin{split}
      \psi(x^{k};x^{k}) & = \mathcal{J}(x^{k}) \\
     \psi^{'}(x^{k};x^{k}) & = \mathcal{J}^{'}(x^{k}).
    \end{split}
\end{equation}
Then $\psi(x;x^{k})$ has the form
\begin{equation}\label{eq:psi}
\begin{split}
  \psi(x;x^{k}) = & \mathcal{J}(x^{k})+(x-x^{k})^{T}\mathcal{J}^{'}(x^{k})\\
    &  +(x-x^{k})^{T}C(x^{k})(x-x^{k})
\end{split}
\end{equation}
with symmetric matrix $C(x^{k})$
\begin{equation}\label{eq:c}
    C(x^{k}) = \sum_{v=1}^{m}\frac{W_{v}^{T}W_{v}}{c^{2}+\|z^{v}-W_{v}x^{k}\|_{2}^{2}}+C_{2}.
\end{equation}

Therefore, we have the following theorem to guarantee the convergence of the IRR algorithm.
\begin{theorem} \label{the:convergence}
The IRR algorithm update in Eq. (\ref{eq:update})  guarantees that
the sequence $\mathcal{J}(x^{k})$ is monotonic, i.e., $\mathcal{J}(x^{k+1})\leq\mathcal{J}(x^{k})$ for all $k$, and the sequence $x^{k}$ will converge to the local minimizer of $\mathcal{J}$.
\end{theorem}

\subsection{View Insufficiency Analysis}

Information theory provides a natural channel to explain the view insufficiency assumption. In particular, for discrete random variables $A, B$ and $C$, the conditional mutual information $I(A;B|C)$ measures how much information is shared between $A$ and $B$ conditioned on already known $C$.

Given an active view set $S$ in the multi-view learning setting, each time we randomly generate view $Z^{v}$ from the latent intact space $\mathcal{X}$,  and add it into $S$. For example, beginning with the view set $S$ already containing one view $Z^{1}$, the insufficiency of the newly generated view $Z^{2}$ can be measured by
\begin{equation}\label{eq:view_inf}
    I(X;Z^{2}|Z^{1})\geq \varepsilon_{info}^{(2)},
\end{equation}
where $\varepsilon_{info}^{(2)}$ is a variable larger than zero. Conventional view sufficiency assumption $I(X;Z^{2}|Z^{1})\leq \varepsilon_{info}$ with some small $\varepsilon_{info}>0$ \cite{sridharan2008information} states that both $Z^{1}$ and $Z^{2}$ are redundant with regards to their information about
$X$. By contrast, our view insufficiency assumption (i.e., Eq. (\ref{eq:view_inf})) implies that an individual view cannot sufficiently describe $X$, and therefore each view will carry additional information about $X$ that other views do not have. For the current view set $S=\{Z^{1}, Z^{2}\}$, the information obtained with respect to $X$ is measured using
\begin{equation}\label{}
    I(X;Z^{1},Z^{2})=I(X;Z^{1})+I(X;Z^{2}|Z^{1}).
\end{equation}
According to the chain rule of mutual information, we have the following proposition.
\begin{proposition}\label{pro:view_inf}
Considering there are $m$ randomly generated views $\{Z^{1}, \cdots, Z^{m}\}$ from the latent comprehensive space $\mathcal{X}$, the information obtained to learn $\mathcal{X}$ can be measured by
\begin{equation}\nonumber
    I(X;Z^{1},\cdots,Z^{m}) = \sum_{i=1}^{m}I(X;Z^{i}|Z^{i-1},\cdots,Z^{1}).
\end{equation}
\end{proposition}
Proposition \ref{pro:view_inf} suggests that more views will bring in more information with respect to $X$. Although each individual view is insufficient, we can receive abundant information to learn the latent intact space $X$ by  exploiting the complementarity between multiple views.

Assume that the latent intact space $\mathcal{X}$ can be completely captured by the ideal view set $S^{*}_{M}=\{Z^{1},\cdots,Z^{M}\}$. Let $X^{*}_{M}$ denote the optimal solution with respect to $S^{*}_{M}$, i.e., $X^{*}_{M}=\min_{X}L(X;S^{*}_{M})$, where $L(\cdot)$ is the expectation of the loss function $\ell(\cdot)$ over the samples on different views. Since $M$ could be very large, we randomly select $m$ views from $S^{*}_{M}$ to construct a smaller view set $S^{*}_{m}$ for multi-view learning. Sridharan and Kakade \cite{sridharan2008information} developed a significant  information theoretical framework to analyze multi-view learning algorithms, based on which we show  the learning error of the newly proposed algorithm is bounded by the following theorem.

\begin{theorem}\label{the:view_inf}
Given the ideal view set $S^{*}_{M}=\{Z^{1},\cdots,Z^{M}\}$, and the bounded loss function $\ell(\cdot)$, the expected losses of multi-view learning based on $S^{*}_{M}$ and its subset $S^{*}_{m}$ are denoted by $L(X;S^{*}_{M})$ and $L(X;S^{*}_{m})$. Their difference is bounded by
\begin{equation}\nonumber
    |L(X;S^{*}_{M})-L(X;S^{*}_{m})|\leq \sqrt{I(X;S^{*}_{M}\backslash S^{*}_{m}|S^{*}_{m})},
\end{equation}
which will decrease with increasing $m$.
\end{theorem}

According to Theorem \ref{the:view_inf}, although we cannot obtain all the necessary views to learn the latent intact space $\mathcal{X}$, we can approximately restore it when provided with enough views.

\subsection{Generalization Error Analysis}

In this section, we propose a new definition of \emph{multi-view stability} and use it with the Rademacher complexity to analyze the generalization error of the proposed multi-view learning algorithm.

In learning theory, algorithmic stability \cite{bousquet2002stability} is employed to measure the variation of the output for different training sets that are identical up to removal or change of a single example. We apply this idea to multi-view learning, and then propose the definition of multi-view stability.

\begin{definition}
Given $f=(f_{1}, \cdots, f_{m})$ and $f\in \mathcal{F}$, where $f_{v}$ is the view function on $v$-view. The function class $\mathcal{F}$ is said to have multi-view stability $\beta$, if for any two multi-view examples $z=\{z^{1}, \cdots, z^{m}\}$ and $\widehat{z}=\{\widehat{z}^{1},  \cdots, \widehat{z}^{m}\}$ that differ only at a single coordinate on an individual view, $\sup_{f \in \mathcal{F}} \|f(z)-f(\widehat{z})\|_{1}\leq \beta$.
\end{definition}

Stability characterizes a system's persistence against the perturbation of variables. On the other hand, since the perturbed variables can be regarded as the outliers (or at least noised examples) for the system, the stability has an implicit connection with robustness. Hence, the theoretical analysis based on the stability here is  expected to deliver similar conclusions from the perspective of robustness.

Conventional concentration inequalities, e.g., Hoeffding's inequality and McDiarmid's inequality, are designed under the independent identical distribution assumption. However, for more complex cases (e.g., multi-view learning),  the variables can have dependence on each other, which calls for the new concentration inequality  developed for dependent random variables.  By leveraging recent results in the concentration of dependent random variables \cite{kontorovich2008concentration}, we use the following concentration inequality for our analysis on multi-view learning.
\begin{theorem}\label{the:con}
Let $\phi: \mathcal{Z}\rightarrow \mathbb{R}$, and $M\in\mathbb{R}^{m\times m}$ be the coefficient matrix measuring variable dependence. For any two multi-view examples $z=\{z^{1}, \cdots, z^{m}\}$ and $\widehat{z}=\{\widehat{z}^{1},  \cdots, \widehat{z}^{m}\}$ that differ only at a single coordinate on an individual view, $|\phi(z)-\phi(\widehat{z})|\leq c$. Then for any $\epsilon>0$,
\begin{equation}\nonumber
\mathbb{P}\{\phi-\mathbb{E}_{z}[\phi] \geq \epsilon \}\leq \exp\left( -\frac{2\epsilon^{2}}{mc^{2}\|M\|_{\infty}^{2}} \right).
\end{equation}
\end{theorem}

Before proceeding, we define some notations for convenience.
Given a multi-view example $z=\{z^{1}, \cdots, z^{m}\}$ and multi-view functions $f=(f_{1}, \cdots, f_{m})$, we define $f(z) = (f_{1}(z^{1}), \cdots, f_{m}(z^{m}))$. For any $f\in \mathcal{F}$, let
\begin{equation}\nonumber
\begin{split}
F(z) = &\frac{1}{m}\sum_{v=1}^{m}f_{v}(z^{v}), \\
\psi(\mathcal{F}, z) = &\sup_{f\in \mathcal{F}} \; [\mathbb{E}_{z}[F(z)]-F(z)].
\end{split}
\end{equation}

Besides multi-view stability, we employ the Rademacher complexity to measure the hypothesis complexity, and its definition is adapted from \cite{bartlett2003rademacher} by removing the assumption that $z^{1}, \cdots, z^{m}$ are i.i.d.

\begin{definition}
Let $z=\{z^{1}, \cdots, z^{m}\}$ be a set of random variables. Let $\{\sigma_{i}\}_{v=1}^{m}$ be a set of independent Rademacher random variables, with zero mean and unit standard deviation. The empirical Rademacher complexity of $\mathcal{F}$ is
\begin{equation}
\mathcal{R}(\mathcal{F},z)   = \mathbb{E}_{\sigma}\left[ \sup_{f\in \mathcal{F}}\frac{1}{m}\sum_{v=1}^{m}\sigma_{v}f_{v}(z^{v})|z\right].
\end{equation}
The Rademacher complexity of $\mathcal{F}$  is $\mathcal{R}_{m}(\mathcal{F}) = \mathbb{E}_{z}[\mathcal{R}(\mathcal{F},z)]$.
\end{definition}

With these definitions, we now present our main result.
\begin{theorem}\label{the:main}
Fix $\delta\in (0,1)$ and $m\geq 1$. If $\mathcal{F}$ has multi-view stability $\beta$, then with probability at least $1-\delta$ over multi-view example $z=\{z^{1}, \cdots, z^{m}\}$,
\begin{equation*}
\sup_{f\in \mathcal{F}} \; \mathbb{E}_{z}[F(z)]-F(z) \leq 2\mathcal{R}_{m}(\mathcal{F}) + \beta \|M\|_{\infty} \sqrt{\frac{\ln(1/\delta)}{2m}}.
\end{equation*}
\end{theorem}

We can directly apply Theorem \ref{the:main} to the setting in which there are $n$ i.i.d. multi-view examples.
\begin{corollary}\label{cor:main}
Fix $\delta\in (0,1)$, $m\geq 1$ and $n \geq 1$. Let $Z=\{z_{i}\}_{i=1}^{n}$ be a set of $n$ multi-view examples. If $\mathcal{F}$ has multi-view stability $\beta$, then with probability at least $1-\delta$ over $Z$,
\begin{equation*}
\sup_{f\in \mathcal{F}} \; \mathbb{E}_{z}[F(Z)]-F(Z) \leq 2\mathcal{R}_{mn}(\mathcal{F}) + \beta \|M\|_{\infty} \sqrt{\frac{\ln(1/\delta)}{2mn}}.
\end{equation*}
\end{corollary}

Through the theoretical analysis, we find that the generalization error of multi-view leaning algorithms can be well bounded by the multi-view stability and Rademacher complexity of the hypothesis. Next we proceed to analyze the specific multi-view stability and Rademacher complexity for the the proposed MISL algorithm.

\subsubsection{Multi-view Stability}

For any  multi-view example $z=\{z^{1},\cdots, z^{m}\}$, we consider that  there is a perturbation   $\tau$ at $j$-th coordinate of the $k$-th view. The new multi-view example is thus written as $\widehat{z}=\{z^{1}, \cdots, z^{k}+\Delta_{k},\cdots, z^{m}\}$, where $\Delta_{k}$ has only one non-zero element $\tau$ at $j$-th coordinate. We suppose $\|W_{v}\|\leq \Omega_{v}$. 

The following proposition illustrates the multi-view stability of the proposed MISL algorithm.
\begin{proposition}\label{pro:stability}
Given $f=(f_{1}, \cdots, f_{m})$ and $f\in \mathcal{F}$, where $f_{v}$ is the view function on $v$-view. The function class $\mathcal{F}$ learned through MISL algorithm has multi-view stability 
\begin{equation}\nonumber
\beta =\frac{\sqrt{2}}{c}|\tau| + \sum_{v=1}^{m} \frac{(128)^{1/4}\Omega_{v}}{c}\sqrt{\frac{|\tau|}{mcC_{2}}}
\end{equation}
\end{proposition}

According to the above analysis, we know that multiple views  together determine the multi-view stability. If one view is perturbed by outliers or noise, the other clean views will act to alleviate the influence of the perturbation and preserve the output results unchanged. This resistance will be further strengthened with the increasing of the number of clean views. Thus, the multi-view learning performance is improved through the corporation between multiple views. However, if we model all views independently, this corporation will be ignored.

\subsubsection{Rademacher Complexity}
 
We proceed to bound the Rademacher complexity of $\mathcal{F}$ by first bounding its covering number \cite{zhou2002covering, zhang2002covering}.
Given $\widetilde{W} = [W_{1};\cdots; W_{m}]$, we assume that $\widetilde{W}$ lies in a sphere with radius $\Upsilon$. The covering number of $\mathcal{F}$ can be bounded by the following lemma.

\begin{lemma}\label{lem:cover}
For $\epsilon>0$ and $\|x\|\leq\chi$, the covering number of $\mathcal{F}$ is upper bounded by
\begin{equation}
\mathcal{N}(\mathcal{F},\epsilon)  \leq (\frac{3\sqrt{2}\chi \Upsilon}{\epsilon c})^{s},
\end{equation}
where the exponent $s$ is the dimension of the constraint set $\widetilde{W}$ in the sense of  its manifold structure.
\end{lemma}

Obviously, $s$ can be upper bounded using $d(D_{1}+\cdots+D_{m})$.
We suggest that if there exist dependencies between multiple views, simultaneously handling multiple views is advantageous over modeling each view independently. In particular, the connections between matrices $\{W_{v}\}_{v=1}^{m}$ will decrease the intrinsic dimension of $\widetilde{W}$, lead to a lower $s$, and then improve the covering number.


By applying the discretization theorem on the covering number, we can easily get the following proposition to bound the Rademacher complexity of the proposed algorithm.
\begin{proposition}\label{pro:gaussian}
The Rademacher complexity of $\mathcal{F}$ learned through MISL algorithm is bounded by
\begin{equation}\nonumber
\mathcal{R}_{m}(\mathcal{F}) \leq \inf_{\epsilon}\sqrt{\frac{2s\ln(3\sqrt{2}\chi\Upsilon/\epsilon c)}{m}} + \epsilon.
\end{equation}
\end{proposition}

Finally, by integrating Propositions \ref{pro:stability} and \ref{pro:gaussian} with Corollary \ref{cor:main}, we can easily get the generalization error bound of the proposed MISL algorithm. Though some views maybe noisy,  multi-view stability can preserve the performance of multi-view learning by exploiting the information on the other accurate views. The dependencies between views can decrease the complexity of the hypotheses space, and then improve the generalization error bound.

\section{Experiments}

In this section, we present qualitative as well as quantitative evaluation on two toys and three real-world datasets. The proposed MISL algorithm and its kernel extension KMISL were compared with convex Multi-view Subspace Learning algorithm (MSL) \cite{white2012convex}, Factorized Latent Spaces with Structured Sparsity algorithm (FLSSS) \cite{jia2010factorized}, and shared Gaussian Latent Variable Model (sGPLVM) \cite{shon2005learning}. 

\subsection{Toy Examples}

First we evaluated our approach on the problem of 3-D point cloud reconstruction. For the 3-D model provided by \cite{wang2012active}, we extracted the point cloud data $X$ (i.e., point positions in rectangular space coordinates), and then projected it into three 2-D planes (e.g., X-Y, X-Z, and Y-Z planes) as the base views $Z^{1}, Z^{2}$ and $Z^{3}$ of $X$. To further validate the robustness of the proposed algorithm, we attempted to generate more noisy views from these base views. Given a randomly placed window of fixed size on each base view, we add some noise to the data points in it, and then obtain a serious of distinct noisy views for multi-view learning.

\begin{figure*}[tb]
\begin{center}
   \includegraphics[width=0.9\textwidth]{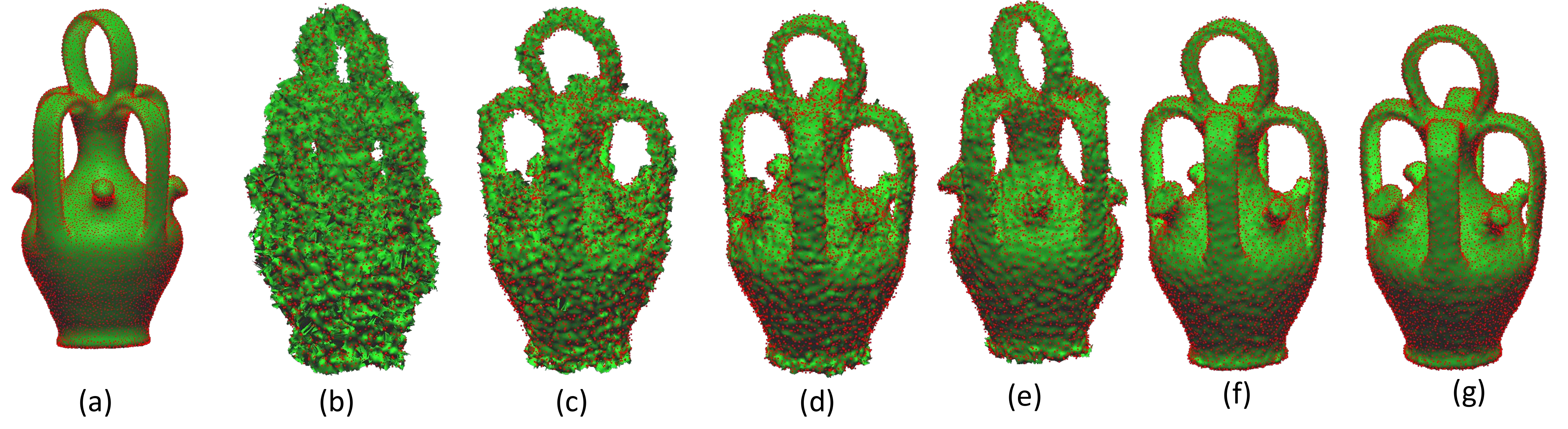}
\end{center}\vskip -0.2in
   \caption{Reconstruction of 3-D vase model using MISL algorithm.} \vskip -0.1in
\label{fig:vase}
\end{figure*}

\begin{figure*}[tb]
\begin{center}
   \includegraphics[width=0.9\textwidth]{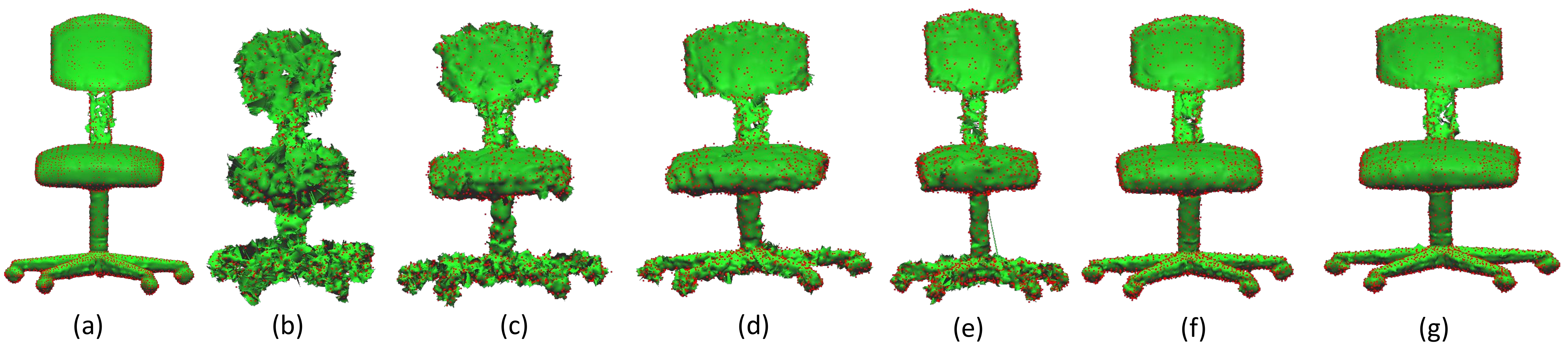}
\end{center}\vskip -0.2in
   \caption{Reconstruction of 3-D chair model using MISL algorithm.}\vskip -0.1in
\label{fig:chair}
\end{figure*}

\begin{figure*}[tb]
\begin{center}
   \includegraphics[width=0.9\textwidth]{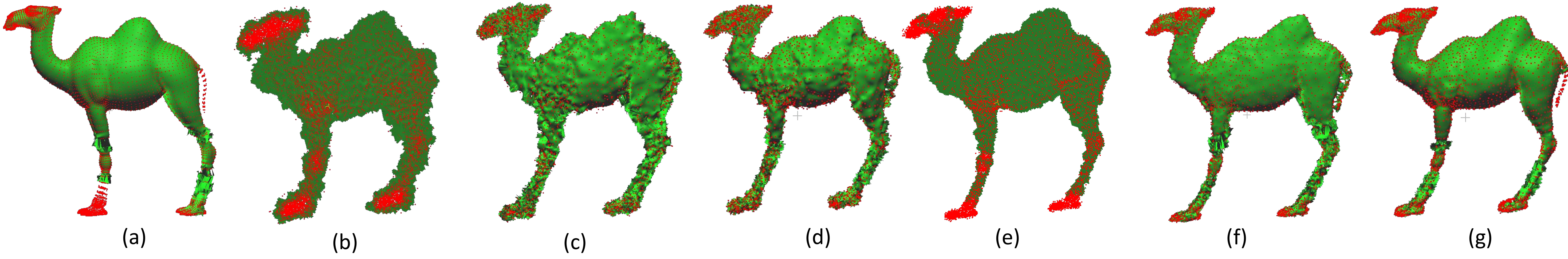}
\end{center}\vskip -0.2in
   \caption{Reconstruction of 3-D compel model using MISL algorithm.}\vskip -0.1in
\label{fig:compel}
\end{figure*}

In Figure \ref{fig:vase}, we show the 3-D latent intact spaces discovered by MISL algorithm with  different numbers of noisy view under distinct signal noise ratio (SNR) as input for vase, chair and compel models, respectively. Specifically, Figure \ref{fig:vase} (a) shows the initial 3-D point cloud in red color and its corresponding meshed result in green. For each base view, we randomly generate 3, 6 and 9 noisy views, and then we totaly have three different settings of 9, 18 and 27 noisy views for training. Given $SNR=10$, we show the intact spaces discovered from these three settings in Figure \ref{fig:vase} (b), (c) and (d), respectively. When $SNR=20$, the intact spaces are presented in Figure \ref{fig:vase} (e), (f) and (d). The reconstruction results under similar settings for chair and compel models are presented in Figures \ref{fig:chair} and \ref{fig:compel}, respectively.
From these reconstruction results, we find that if the signal is  cleaner (i.e.,  associated with a higher SNR), the reconstructed object will be more accurate. More importantly, although each individual view is noisy and insufficient for completely inferring the 3-D objects, MISL algorithm is robust enough to outliers and can accurately discover the intact space by utilizing the complementarity between these views.

\begin{figure*}[tb]
\begin{center}
   \includegraphics[width=0.8\textwidth]{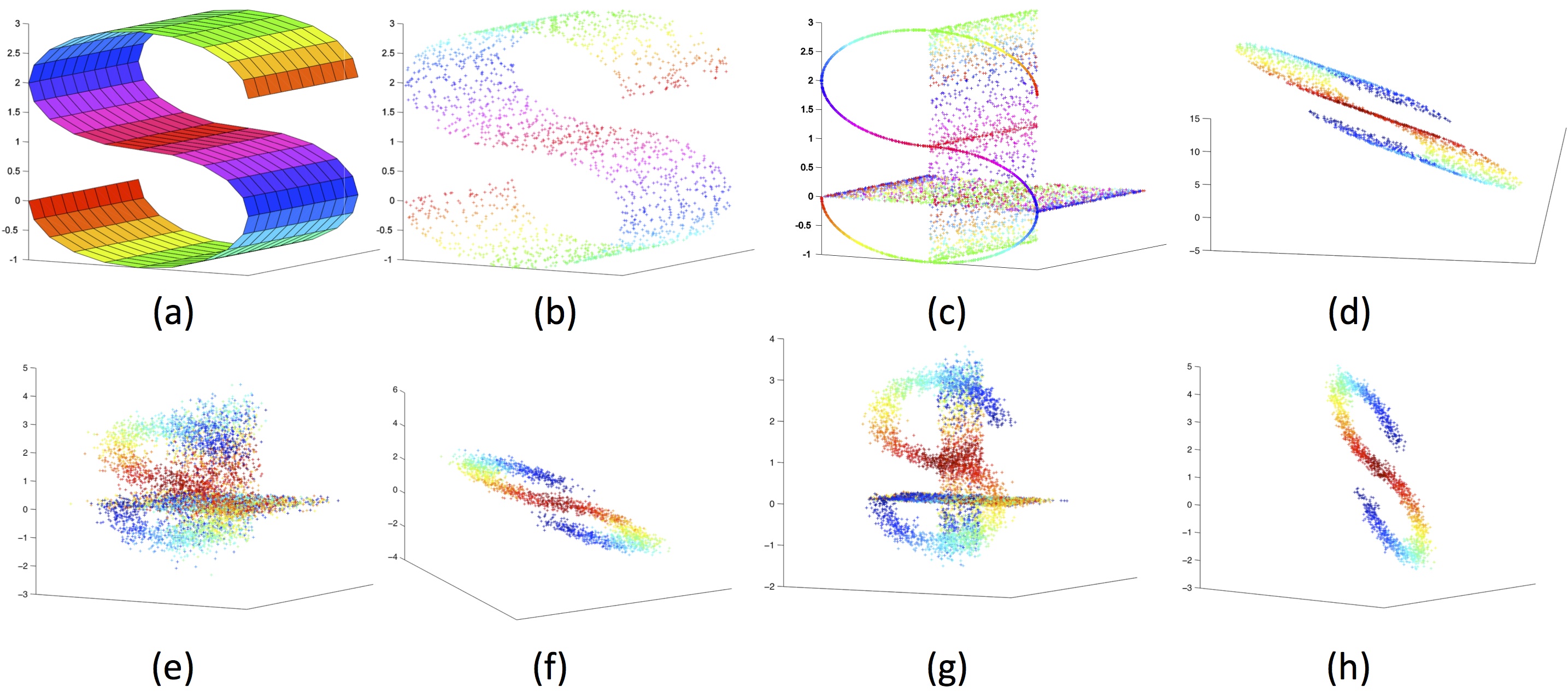}
\end{center}\vskip -0.1in
   \caption{Reconstruction of s-curve using MISL algorithm.} \vskip -0.25in
\label{fig:s_curve}
\end{figure*}

The second toy is based on the synthetic data ``S-curve'' as shown in Figure \ref{fig:s_curve} (a). We uniformly sampled some 3-dimensional data points $X$ (see Figure \ref{fig:s_curve} (b)), and then projected them into three 2-D planes (e.g., X-Y, X-Z, and Y-Z planes) as three views $Z^{1}, Z^{2}$ and $Z^{3}$ of $X$ (see Figure \ref{fig:s_curve} (c)). Based on these three  views, we find that the proposed MISL algorithm can effectively reconstruct the s-curve, as shown in Figure \ref{fig:s_curve} (d).
We further add some noises on the clean views  to evaluate the robustness of MISL. Figures \ref{fig:s_curve} (e) and (g) show the noisy views of $SNR=15$ and $SNR=20$, respectively. Specifically, it has already became difficult for us human being to figure out the original curve from the terrible views in Figure \ref{fig:s_curve} (e).  Thanks to the robustness of MISL, the noise can be appropriately handled, and the intact spaces (see Figures \ref{fig:s_curve} (f) and (h)) under $SNR=15$ and $SNR=20$ will be approximately restored.

\subsection{Face Recognition}

\begin{figure*}[tb]
\begin{center}
   \includegraphics[width=0.9\textwidth]{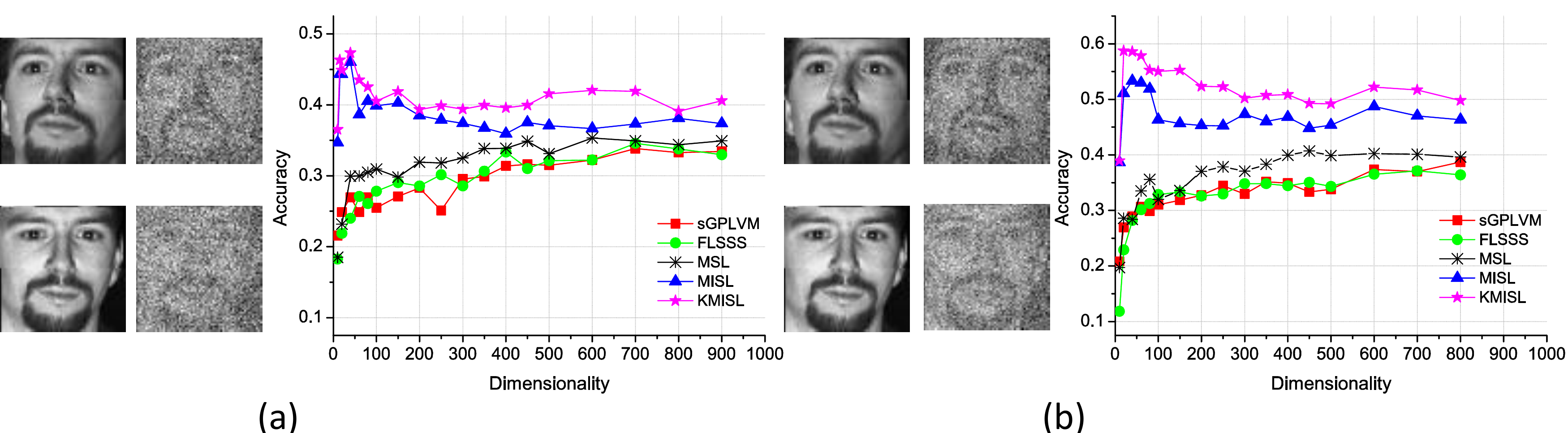}
\end{center}\vskip -0.1in
   \caption{Face recognition accuracies of different algorithms on different dimensional spaces.}\vskip -0.1in
\label{fig:face}
\end{figure*}

CMU PIE face database \cite{sim2002cmu} contains 41,368 images of 68 people, each person under 13 different poses, 43 different illumination conditions, and with 4 different expressions. To construct the multi-view setting, we selected two near frontal poses (i.e., C9 and C29) as two views. Therefore a pair of images of one person under these two poses with the same illumination can be seen as a two-view example. Different algorithms were used to project the multi-view faces into some appropriate spaces for face recognition. $50\%$ images of one people were randomly selected for training, and the rest for test. $k$-nearest neighbor method based on the Euclidean distance was applied for face recognition, where $k$ was set as 3. Given the noisy views of $SNR=2$ and $SNR=5$, the face recognition accuracies for different algorithms on different dimensional spaces were shown in Figure \ref{fig:face} (a) and (b).

\begin{figure}[tb]
\begin{center}
   \includegraphics[width=0.8\columnwidth]{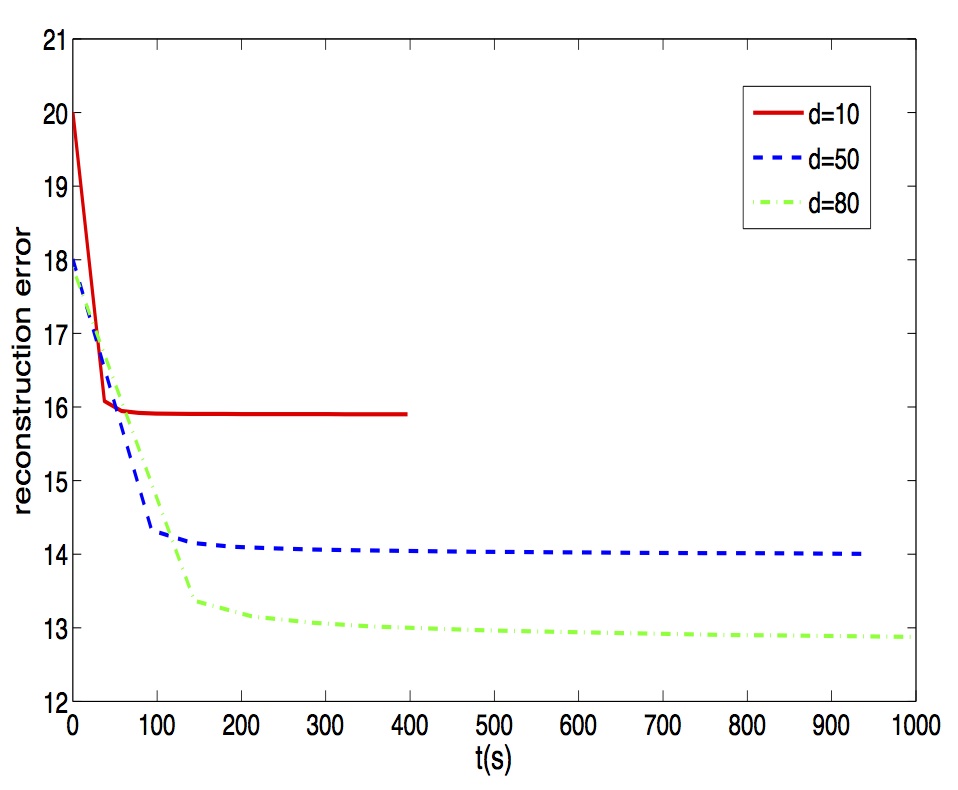}
\end{center}\vskip -0.15in
   \caption{Convergence curves of MISL for different dimensional latent  spaces.}\vskip -0.25in
\label{fig:convergence}
\end{figure}

From Figure \ref{fig:face}, we find that MISL stably outperforms other algorithms at all the dimensionalities. The noisy views do not seriously damage the performance of MISL, but are optimally combined to find the latent intact space. This is mainly due to the satisfied robust property of MISL and its ability to appropriately handle the complementarity between multiple views.

In Figure \ref{fig:convergence}, we present the convergence curves of MISL algorithm in discovering different low-dimensional intact spaces. It is instructive to note that IRR is an effective method to optimize MISL, and leads to a fast convergence of the reconstruction error (i.e., Eq. (\ref{eq:recon})).  
\vspace {-0.15in}
\subsection{Human Motion Recognition}


The UCF101 dataset \cite{soomro2012ucf101} is a large dataset of human actions. It consists of 101 action classes, which can be further divided into five types: Human-Object Interaction, Body-Motion Only, Human-Human Interaction, Playing Musical Instrument and Sports. There are totally over 13k clips and 27 hours of video data. The entire dataset was split between train and test samples three times, each split randomly selecting two-thirds of the data for training and the remaining data for testing. Therefore videos from the same group never appear in both the training and test set. 
Recently, deep learning models \cite{ji20133d,ning2005toward} have been widely used to learn effective features from videos for action recognition. Since feature learning is not in our research interests, we employ the easily obtained conventional handcrafted features as input, and launch multi-view learning to evaluate whether the complementarity between multiple views can actually been exploited to improve the recognition performance.
Motion Boundary Histograms (MBH)  and Histograms of Oriented Gradients (HOG) are two well-known motion descriptors \cite{wang2013action}, and thus we chose these two views to represent each clip. In the experiments, we used a linear SVM for classification, and conduct cross validation to find the optimal $C$. In the case of multi-class classification, we used a one-against-rest approach and select the class with the highest score. The performance measurement was the average classification accuracy over all classes on three splits.

\begin{figure*}[tb]
\begin{center}
   \includegraphics[width=0.8\textwidth]{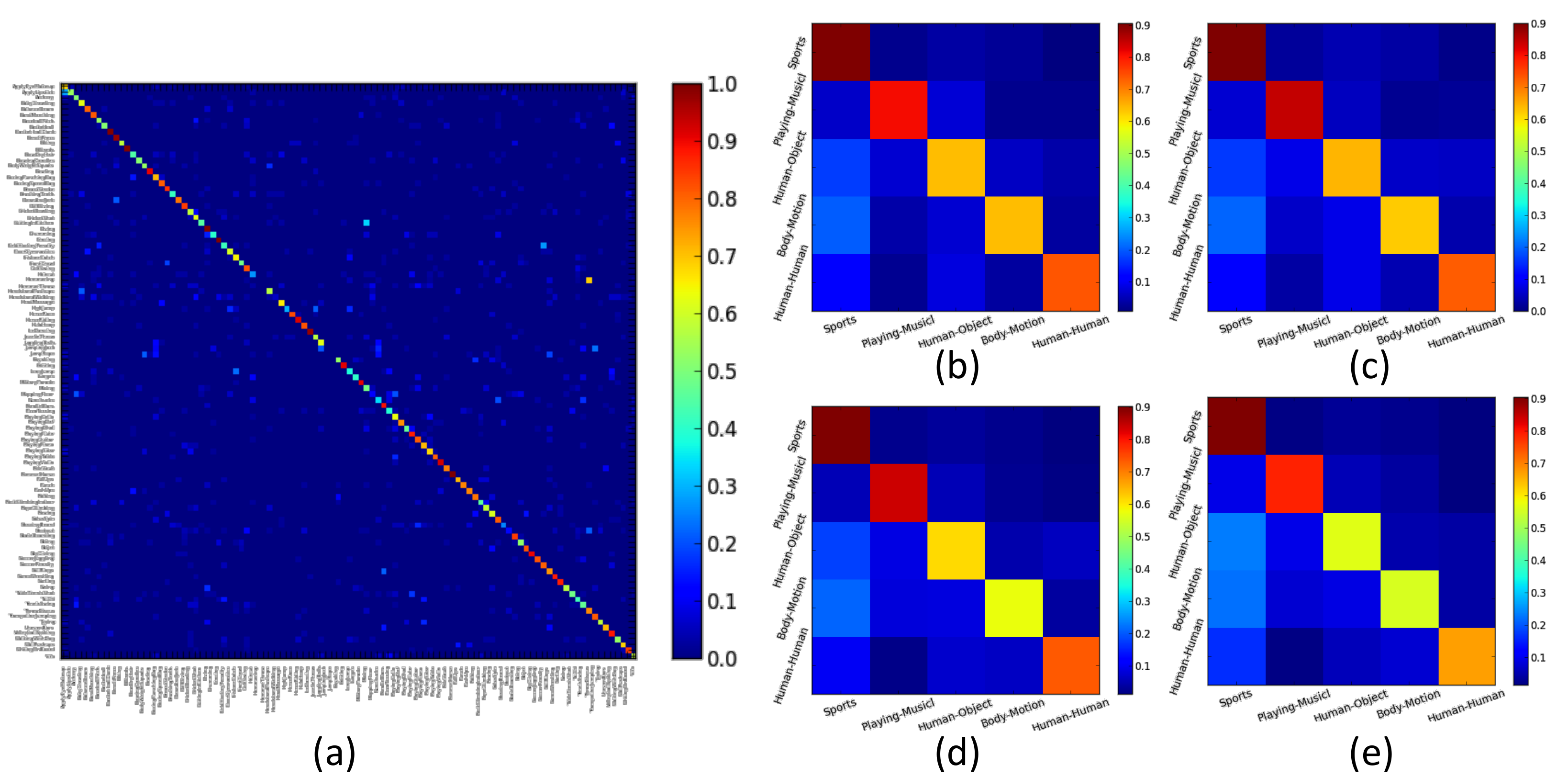}
\end{center}\vskip -0.15in
   \caption{Confusion tables of different algorithms on 101 action classes and 5 action types.}\vskip -0.2in
\label{fig:motion_conf}
\end{figure*}

\begin{table}[tb]
\caption{Average performance of different algorithms.}\vskip -0.2in
\label{tab:motion}
\begin{center}

{\small
\begin{tabular}{lcccc}
\hline
MBH/HOG & MBH+HOG & MBH & HOG \\
\hline
sGPLVM    & $58.43\%$ & $52.80\%$&$42.02\%$ \\
FLSSS    & $60.75\%$ & $53.56\%$& $43.12\%$ \\
MSL    & $60.49\%$ & $54.70\%$& $42.74\%$ \\
MISL   & $62.57\%$ & $55.75\%$ & $43.81\%$ \\
KMISL   & $64.23\%$ & $56.23\%$ & $44.62\%$ \\
\hline
\end{tabular}
}
\end{center}
\vskip -0.4in
\end{table}

We utilized different algorithms to project the multi-view video clips into the 1500-dimensional latent spaces and reported the recognition accuracy based on these embedded examples. We presented the confusion matric for KMISL over 101 action classes in Figure \ref{fig:motion_conf} (a). After merging the recognition results of classes belonging to the same action types, the confusion matrices over  five action types for MISL, MSL, FLSSS, and sGPLVM algorithms are shown in Figure \ref{fig:motion_conf} (b)-(e) respectively. MISL provides performance improvements in most of the action types.  The recognition accuracy of different algorithms based on various feature combinations is summarized in Table \ref{tab:motion}. Compared with the recognition performance based on single kind of feature as input, learning with multiple features through different multi-view learning algorithms demonstrate variable performance improvements, as a result of the exploitation of complementary information underlying the multi-view features. In particular, MISL obtains the best recognition result, due to its robust approach to obtain the intact latent space.
\vspace {-0.1in}
\subsection{RGB-D Object Recognition}

RGB-D datset is a large-scale multi-view object dataset collected through an RGB-D camera. This dataset contains color and depth images of 300 physically distinct everyday objects. Therefore, the RGB image and the corresponding depth image can be seen as two approaches capturing the shape and the visual appearance of an object from 51 categories. Video sequences of each object were recorded at 20 Hz, and then subsampled by taking every fifth video frame, giving 41,877 RGB and depth images for the experiments. We extracted 1000-dimensional gradient kernel descriptors to depict both the RGB image and depth image. For category recognition, we randomly left one object out from each category for testing and training SVM classifiers on all the remaining objects. The performance were measured in averaged recognition accuracy across 10 trials.

\begin{figure}[tb]
\begin{center}
   \includegraphics[width=0.8\columnwidth]{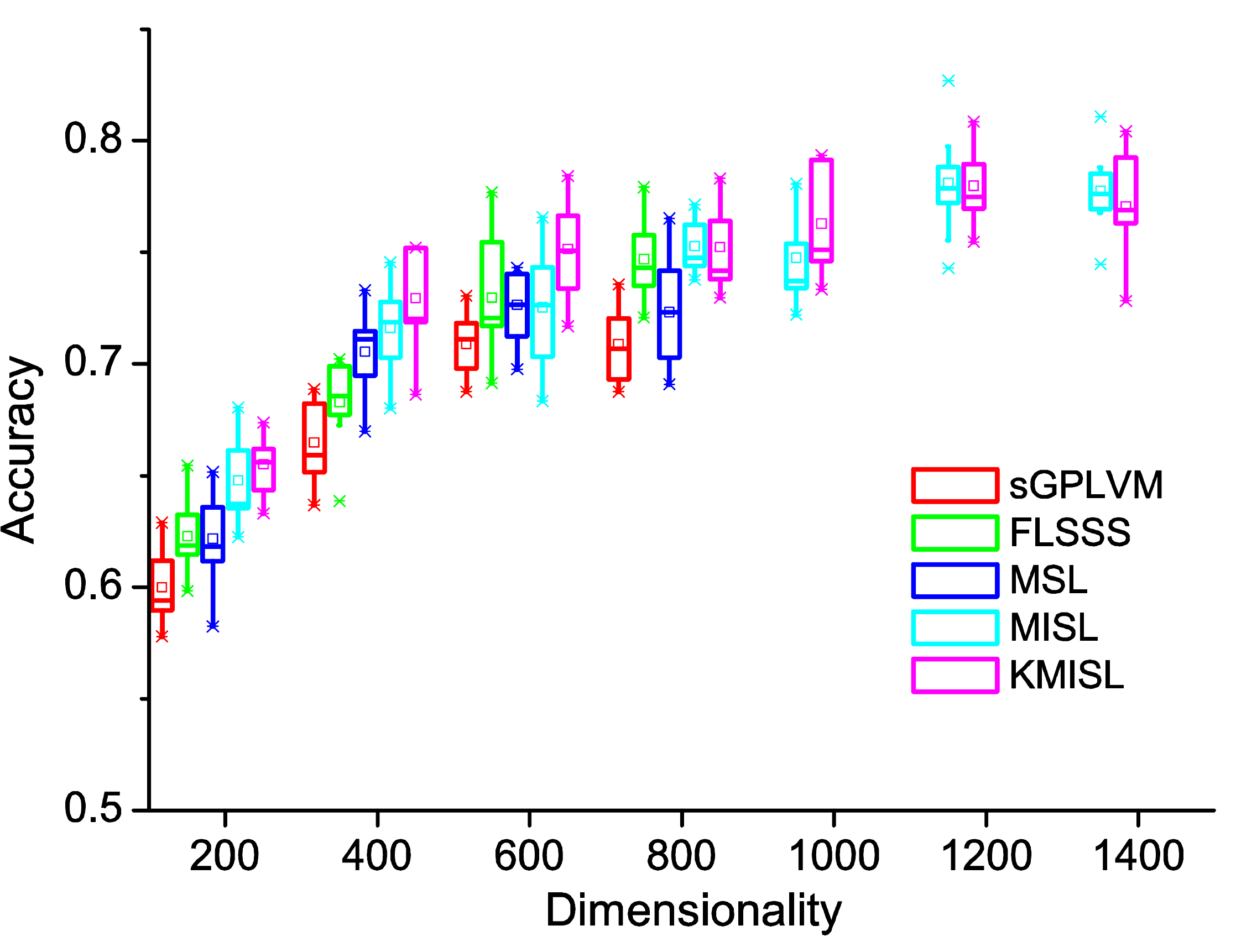}
\end{center}\vskip -0.15in
   \caption{Category recognition accuracy of different algorithms on different dimensional spaces.}\vskip -0.26in
\label{fig:rgbd_acc}
\end{figure}

We used different algorithms to project the multi-view objects into different dimensional spaces and reported the category recognition accuracy on these spaces in Figure \ref{fig:rgbd_acc}. We can see that both MISL and its kernel extension perform much better than other competitors. In contrast with these competitors, whose objective is to discover the subspace bearing dependencies or independencies of different views, the proposed MISL algorithm attempts to find an intact space fusing all the information provided by diverse views. Therefore the dimensionality of the spaces discovered by MISL is not limited to be lower than those of original features. Moreover, the robustness of MISL ensures that the noises introduced by multiple views can be appropriately handled, which leads to the performance improvement of multi-view learning.
\vspace {-0.12in}
\section{Conclusion}

We have presented a novel robust learning algorithm for recovering the latent intact representations for multi-view examples, by assuming view insufficiency, i.e., that each view only captures partial information but all views together carry redundant information about the object. Theoretical analysis on view insufficiency assumption suggests that we can approximately restore the latent intact space by exploiting the complementarity between multiple insufficient views. Based on a new definition of ``multi-view stability'' and the Rademacher complexity, we derive the generalization error bound for the proposed multi-view learning algorithm, and show that this bound can be improved  through the complementarity between multiple views. Finally, we design a new Iteratively Reweight Residuals (IRR) technique that converges fast to solve the optimization problem. Experimental results on synthetic data and real-world datasets demonstrate that the proposed MISL algorithm and its kernel extension are robust, effective and promising for practical applications.
\vspace {-0.1in}
\section{Proofs}

We provide below the detailed proofs of the theoretical results in Section 4.
\vspace {-0.1in}
\subsection{Proof of Theorem \ref{the:convergence}}

Assume that $\psi(x;x^{k})$ is locally convex with respect to $x$ and has a local minimizer. Setting $x^{k+1}$ as this minimizer, we have
\begin{equation}\label{eq:10}
    \psi^{'}(x^{k+1};x^{k})= \mathcal{J}^{'}(x^{k})+2C(x^{k})(x^{k+1}-x^{k})=0.
\end{equation}
Substituting for $\mathcal{J}^{'}$, we obtain the update rule in Eq. (\ref{eq:update}) (similar for Eq. (\ref{eq:update_2})).

By appropriately choosing $x^{k}$ near $x$, we have  $\mathcal{J}(x)\leq\psi({x;x^{k}})$ implying that
\begin{equation}\label{eq:11}
    \begin{split}
      \mathcal{J}(x^{k+1}) & \leq \psi(x^{k+1};x^{k}) \\
        & = \mathcal{J}(x^{k})+(x^{k+1}-x^{k})^{T}\mathcal{J}^{'}(x^{k})\\
        &+(x^{k+1}-x^{k})^{T}C(x^{k})(x^{k+1}-x^{k}).
    \end{split}
\end{equation}
By combining Eqs. (\ref{eq:10}) and (\ref{eq:11}), we have
{\small
\begin{equation}\label{eq:12}
    \begin{split}
     \mathcal{J}(x^{k+1})-\mathcal{J}(x^{k})&\leq -(x^{k+1}-x^{k})^{T}C(x^{k})(x^{k+1}-x^{k})\\
        & \leq -C_{2}\|x^{k+1}-x^{k}\|^{2} \leq 0.
    \end{split}
\end{equation}
}\noindent
This shows that $\mathcal{J}(x^{k+1})<\mathcal{J}(x^{k})$, proving the first part of Theorem \ref{the:convergence}.

Since the sequence $\mathcal{J}(x^{k})$ is monotonic and lower bounded, it converges. From Eq. (\ref{eq:12}), we can then write
\begin{equation}\label{}
    \|x^{k+1}-x^{k}\|^{2} \leq \frac{1}{C_{2}}[\mathcal{J}(x^{k})-\mathcal{J}(x^{k+1})].
\end{equation}
Considering $(\mathcal{J}(x^{k}))$ is convergent, we conclude that $(x^{k})$ has convergent subsequences.

In non-convex optimization \cite{singer2007duality}, a common assumption is that the non-convex function $\mathcal{J}(\cdot)$ is ``locally convex'' around its ``local'' minimum. Suppose we assume that the initialization $x^{0}$ is ``good'', in that it is situated sufficiently close a local minimizer $x^{*}$. It is then possible that the entire $(x^{k})$ is restricted to a ball $B_{r}(x^{*})$ of small enough radius $r$ around $x^{*}$, and the ball contains no other stationary points of $\mathcal{J}$. In this case, we are guaranteed that every convergent subsequence, and hence the whole sequence $(x^{k})$, converges to $x^{*}$.

\vspace {-0.2in}
\subsection{Proof of Theorem \ref{the:view_inf}}

Considering $|\ell(x)|\leq 1$, and two probability distributions $P$ and $Q$, we have \begin{equation}\label{}
    \begin{split}
      |\int \ell(x)dQ & -\int\ell(x)dP| = |\int(1-\beta)\ell dQ| \\
        & \leq \int |1-\beta|dQ \leq \sqrt{KL(Q;P)},\\
    \end{split}
\end{equation}
where $\beta=\frac{dP}{dQ}$ and the last step is due to the $L_{1}$ variational distance being bounded by the square root of the KL divergence. Using this, for a fixed view set we have
\begin{equation}\label{}
    \begin{split}
      &|E_{X|S^{*}_{m}}\ell(X;S^{*}_{M})-E_{X|S^{*}_{M}}\ell(X;S^{*}_{M})|  \\
        & \leq \sqrt{KL(P_{X|S^{*}_{M}};P_{X|S^{*}_{m}})}.
    \end{split}
\end{equation}
Taking the expectation with respect to $S^{*}_{M}$ and using Jensen's inequality, we get
\begin{equation}\label{}
    \begin{split}
      &|E_{S^{*}_{M}}E_{X|S^{*}_{m}}\ell(X;S^{*}_{M})-L(X;S^{*}_{M})|  \\
        & \leq \sqrt{E_{S^{*}_{M}}KL(P_{X|S^{*}_{M}};P_{X|S^{*}_{m}})}.
    \end{split}
\end{equation}
Since
\begin{equation}\label{}
    L(X;S^{*}_{m})\leq E_{S^{*}_{M}}E_{X|S^{*}_{m}}\ell(X;S^{*}_{M}),
\end{equation}
and  $L(X;S^{*}_{M}) \leq L(X;S^{*}_{m})$, we obtain
{\small
\begin{equation}\label{}
    |L(X;S^{*}_{M})-L(X;S^{*}_{m})|\leq \sqrt{E_{S^{*}_{M}}KL(P_{X|S^{*}_{M}};P_{X|S^{*}_{m}})}.
\end{equation}
}\noindent
Considering
{
\begin{equation}\label{}
    E_{S^{*}_{M}}KL(P_{X|S^{*}_{M}};P_{X|S^{*}_{m}}) = I(X;S^{*}_{M}\backslash S^{*}_{m}|S^{*}_{m}),
\end{equation}
}\noindent
we have
{
\begin{equation}\label{}
     |L(X;S^{*}_{M})-L(X;S^{*}_{m})|\leq \sqrt{I(X;S^{*}_{M}\backslash S^{*}_{m}|S^{*}_{m})}.
\end{equation}
}\noindent
Furthermore,
\begin{equation}\nonumber
    \begin{split}
      I&(X;S^{*}_{M}\backslash S^{*}_{m-1}|S^{*}_{m-1})-I(X;S^{*}_{M}\backslash S^{*}_{m}|S^{*}_{m})\\
        & = H(X,S^{*}_{m-1})+H(S^{*}_{M})-H(X,S^{*}_{M})-H(S^{*}_{m-1})\\
        & -\left(H(X,S^{*}_{m})+H(S^{*}_{M})-H(X,S^{*}_{M})-H(S^{*}_{m})\right)\\
        & = H(Z_{m}|S^{*}_{m-1})-H(Z_{m}|X,S^{*}_{m-1})\\
        & = I(X,Z_{m}|S^{*}_{m-1})\geq 0,
    \end{split}
\end{equation}
which shows that the larger $m$, the less $I(X;S^{*}_{M}\backslash S^{*}_{m}|S^{*}_{m})$ will be, and the difference between $L(X;S^{*}_{M})$ and $L(X;S^{*}_{m})$ will decrease.

\vspace {-0.2in}
\subsection{Proof of Theorem \ref{the:main}}

For any two multi-view examples $z=\{z^{1}, \cdots, z^{m}\}$ and $\widehat{z}=\{\widehat{z}^{1},  \cdots, \widehat{z}^{m}\}$ that differ only at a single coordinate on an individual view, assume that $\psi(\mathcal{F}, z)\geq \psi(\mathcal{F}, \widehat{z})$. We have
\begin{equation}\nonumber
\begin{split}
|\psi&(\mathcal{F}, z) -  \psi(\mathcal{F}, \widehat{z})|\\
& = \left|\sup_{f\in \mathcal{F}} \; [\mathbb{E}_{z}[F(z)]-F(z)] - \sup_{\widehat{f}\in \mathcal{F}} \; [\mathbb{E}_{\widehat{z}}[\widehat{F}(\widehat{z})]-\widehat{F}(\widehat{z})]\right| \\
& \leq \left|\sup_{f\in \mathcal{F}} \; [\mathbb{E}_{z}[F(z)]-F(z)-\mathbb{E}_{\widehat{z}}[F(\widehat{z})]+ F(\widehat{z})] \right|\\
& = \left|\sup_{f\in \mathcal{F}} \; [\frac{1}{m}\sum_{v=1}^{m}f_{v}(\widehat{z}^{v})-f_{v}(z^{v})]\right|\\
& \leq \sup_{f\in \mathcal{F}} \; \frac{1}{m}\|f(\widehat{z})-f(z)\|_{1}\leq \frac{\beta}{m}.
\end{split}
\end{equation}
The last equality results from the multi-view stability. According to Theorem \ref{the:con}, we therefore have
\begin{equation}
\psi(\mathcal{F}, z) \leq \mathbb{E}_{z}[\psi(\mathcal{F}, z)] + \beta \|M\|_{\infty} \sqrt{\frac{\ln(1/\delta)}{2m}},
\end{equation}
with probability at least $1-\delta$.

Using the symmetry argument and the contraction principle, we proceed to upper-bound $\mathbb{E}_{z}[\psi(\mathcal{F}, z)] $ by the Rademacher complexity $\mathcal{R}_{m}(\mathcal{F})$. It is necessary to note that our analysis does not require the random variables to be independent.

Based on  Jensen's inequality, we begin with the definition of $\mathbb{E}_{z}[\psi(\mathcal{F}, z)] $ and have
\begin{equation}\nonumber
\begin{split}
\mathbb{E}_{z}[\psi(\mathcal{F}, z)]  =& \mathbb{E}_{z} \left[ \sup_{f\in \mathcal{F}} \; [\mathbb{E}_{z}[F(\widehat{z})]-F(z)] \right]\\
\leq & \mathbb{E}_{z} \left[ \sup_{f\in \mathcal{F}} \; [F(\widehat{z})-F(z)] \right].
\end{split}
\end{equation}
Given a set of independent variables $\{\sigma_{v}\}_{v=1}^{m}$, uniformly distributed on $\{-1, 1\}$, we define
\begin{equation}
h_{\sigma_{v}}(z,\widehat{z})\left\{
  \begin{aligned}
  z & \quad \texttt{if} \; \sigma_{v}=1,\\
  \widehat{z} & \quad  \texttt{if} \; \sigma_{v} = -1,
  \end{aligned}
\right.
\end{equation}
and
\begin{equation}
\widehat{h}_{\sigma_{v}}(z,\widehat{z})\left\{
  \begin{aligned}
  \widehat{z} & \quad \texttt{if} \; \sigma_{v}=1,\\
  z & \quad  \texttt{if} \; \sigma_{v} = -1.
  \end{aligned}
\right.
\end{equation}
By symmetry,
{\small
\begin{equation}\nonumber
\begin{split}
\mathbb{E}_{z}&[\psi(\mathcal{F}, z)] \leq  \mathbb{E}_{z}  \left[ \sup_{f\in \mathcal{F}} \; \frac{1}{m}\sum_{v=1}^{m} f_{v}(\widehat{z}^{v})-f_{v}(z^{v})\right] \\
= & \mathbb{E}_{\sigma}  \left[ \mathbb{E}_{z}  \left[  \sup_{f\in \mathcal{F}} \; \frac{1}{m}\sum_{v=1}^{m} f_{v}(\widehat{h}_{\sigma_{v}}(z^{v},\widehat{z}^{v}))-f_{v}(h_{\sigma_{v}}(z^{v},\widehat{z}^{v})) | \sigma\right] \right] \\
= & \mathbb{E}_{\sigma, z}  \left[ \sup_{f\in \mathcal{F}} \; \frac{1}{m}\sum_{v=1}^{m} \sigma_{v}(f_{v}(\widehat{z}^{v})-f_{v}(z^{v}))\right] \\
\leq & 2\mathbb{E}_{\sigma, z}  \left[ \sup_{f\in \mathcal{F}} \; \frac{1}{m}\sum_{v=1}^{m} \sigma_{v}f_{v}(z^{v}) \right]=2\mathcal{R}_{m}(\mathcal{F}) .
\end{split}
\end{equation}
}\noindent
Finally, we can complete the proof by combining the above results.

\vspace {-0.2in}
\subsection{Proof of Proposition \ref{pro:stability}}

We obtain the  optimal intact representations of multi-view examples $z$ and $\widehat{z}$ through
{\small
\begin{equation}\label{eq:f_x}
\min_{x} \;  \frac{1}{m}\sum_{v=1}^{m} \log\left (1+\frac{\|z^{v}-W_{v}x\|^2}{c^2}\right) + C_{2}\|x\|^{2}
\end{equation}
}\noindent
and
{\small
\begin{equation}\label{eq:f_xhat}
\min_{\widehat{x}} \;  \frac{1}{m}\sum_{v=1}^{m} \log\left (1+\frac{\|\widehat{z}^{v}-W_{v}\widehat{x}\|^2}{c^2}\right) + C_{2}\|\widehat{x}\|^{2},
\end{equation}
}\noindent
and  define $\Delta_{x} = \widehat{x}  - x$.
We introduce the following notation for convenience
\begin{equation}
R(z, x) = \frac{1}{m}\sum_{v=1}^{m} \log\left (1+\frac{\|z^{v}-W_{v}x\|^2}{c^2}\right).
\end{equation}
Though $R(z, x)$ is not globally convex w.r.t. $x$, it can be assumed to be locally convex within a  small region. Given $z$ and its perturbated copy $\widehat{z}$, we assume that their intact representations $x$ and $\widehat{x}$ are not far away from each other, and fulfill the following inequalities derived from the  convex principle that $\forall t\in [0,1]$
\begin{equation}
 R(\widehat{z}, x+t\Delta_{x}) -  R(\widehat{z}, x) \leq t\big(R(\widehat{z}, \widehat{x})-R(\widehat{z}, x)\big)
\end{equation}
and
\begin{equation}
 R(\widehat{z}, \widehat{x}-t\Delta_{x}) -  R(\widehat{z}, \widehat{x}) \leq t\big(R(\widehat{z}, x)-R(\widehat{z}, \widehat{x})\big).
\end{equation}
Summing the two inequalities yields
{\small
\begin{equation}\label{eq:4r}
 R(\widehat{z}, x+t\Delta_{x}) -  R(\widehat{z}, x) +  R(\widehat{z}, \widehat{x}-t\Delta_{x}) -  R(\widehat{z}, \widehat{x})  \leq 0.
\end{equation}
}\noindent
Considering that $x$ and $\widehat{x}$ are optimal solutions of Eq. (\ref{eq:f_x}) and Eq. (\ref{eq:f_xhat}), respectively, we have
\begin{equation}\label{eq:opt1}
R(z, x) + C_{2}\|x\|^{2} \leq R(z, x + t\Delta_{x}) + C_{2}\|x+ t\Delta_{x}\|^{2} 
\end{equation}
and
\begin{equation}\label{eq:opt2}
R(\widehat{z}, \widehat{x}) + C_{2}\|\widehat{x}\|^{2} \leq R(\widehat{z}, \widehat{x} - t\Delta_{x}) + C_{2}\|\widehat{x}- t\Delta_{x}\|^{2} 
\end{equation}
Using Eq. (\ref{eq:4r}),  Eq. (\ref{eq:opt2}) can be rewritten as
\begin{equation}\label{eq:opt22}
 R(\widehat{z}, x+t\Delta_{x}) -  R(\widehat{z}, x) \leq C_{2}\|\widehat{x}- t\Delta_{x}\|^{2} - C_{2}\|\widehat{x}\|^{2}.
\end{equation}
Summing Eqs. (\ref{eq:opt1}) and (\ref{eq:opt22}), we obtain
{\scriptsize
\begin{equation}\nonumber
\begin{split}
&mC_{2}\bigg(\|x \|^{2} -\|x+ t\Delta_{x}\|^{2} - \|\widehat{x}- t\Delta_{x}\|^{2} + \|\widehat{x}\|^{2} \bigg)\\
\leq & \log(1+\frac{\|z^{k}-W_{v}(x+t\Delta_{x})\|^2}{c^2}) - \log(1+\frac{\|\widehat{z}^{k}-W_{v}(x+t\Delta_{x})\|^2}{c^2}) + \\
& \log(1+\frac{\|\widehat{z}^{k}-W_{v}x\|^2}{c^2}) - \log(1+\frac{\|z^{k}-W_{v}x\|^2}{c^2}) \\
\leq & \frac{\sqrt{2}}{c} \bigg(\|z^{k}-W_{v}(x+t\Delta_{x})\| - \|\widehat{z}^{k}-W_{v}(x+t\Delta_{x})\| + \\
& \quad \quad  \quad \|\widehat{z}^{k}-W_{v}x\|- \|z^{k}-W_{v}x\|\bigg) \\
\leq & \frac{2\sqrt{2}}{c}\|z^{k} - \widehat{z}^{k}\| = \frac{2\sqrt{2}|\tau|}{c}
\end{split}
\end{equation}
}\noindent

By setting $t=\frac{1}{2}$, the above inequality can then be reformulated as
\begin{equation}
\|\Delta_{x}\|^{2} \leq \frac{4\sqrt{2}|\tau|}{mcC_{2}}
\end{equation}
For any view except $k$-th view, the difference between $f_v(z^{v},x)$ and $f_v(z^{v},\widehat{x})$ is 
{\small
\begin{equation}\nonumber
\begin{split}
|f_{v}&(z^{v},x)  - f_{v}(z^{v},\widehat{x}) |\\
 = & \left| \log\left (1+\frac{\|z^{v}-W_{v}x\|^2}{c^2}\right) - \log\left (1+\frac{\|z^{v}-W_{v}\widehat{x}\|^2}{c^2}\right) \right|\\
 \leq & \frac{\sqrt{2}}{c} \bigg| \|z^{v}-W_{v}x\| - \|z^{v}-W_{v}\widehat{x}\|\bigg| \\
 \leq & \frac{\sqrt{2}}{c}\|W_{v}\Delta_{x}\| \leq \frac{(128)^{1/4}\Omega_{v}}{c}\sqrt{\frac{|\tau|}{mcC_{2}}}
\end{split}
\end{equation}
}\noindent
For the $k$-th view, we have
{\small
\begin{equation}\nonumber
\begin{split}
|f_{k}&(z^{k}, x)  - f_{k}(\widehat{z}^{k}, \widehat{x}) |\\
 = & \left| \log\left (1+\frac{\|z^{k}-W_{k}x\|^2}{c^2}\right) - \log\left (1+\frac{\|\widehat{z}^{k}-W_{k}\widehat{x}\|^2}{c^2}\right) \right|\\
 \leq & \frac{\sqrt{2}}{c} \bigg| \|z^{k}-W_{k}x\| - \|\widehat{z}^{k}-W_{k}\widehat{x}\|\bigg| \\
 \leq & \frac{\sqrt{2}}{c} \bigg(\|z^{k} - \widehat{z}^{k}\| + \|W_{k}\Delta_{x}\| \bigg) \\
 \leq & \frac{(128)^{1/4}\Omega_{v}}{c}\sqrt{\frac{|\tau|}{mcC_{2}}} + \frac{\sqrt{2}}{c}|\tau|
\end{split}
\end{equation}
}\noindent
Therefore, $\beta=\|f(z) - f(\widehat{z}) \|_{1}$ can be bounded by
{\small
\begin{equation}\nonumber
\begin{split}
\|f(z) - f(\widehat{z}) \|_{1} \leq &  \frac{\sqrt{2}}{c}|\tau| + \sum_{v=1}^{m} \frac{(128)^{1/4}\Omega_{v}}{c}\sqrt{\frac{|\tau|}{mcC_{2}}}
\end{split}
\end{equation}
}\noindent
\vspace {-0.35in}
\subsection{Proof of Lemma \ref{lem:cover}}

Let us consider the function class $\mathcal{F}_{v}=\{f_{v}(g_{v}(\cdot)): g_{v}\in \mathcal{G}_{v}\}$ on view-$v$. In the proposed algorithm, $f_{v}(\cdot)$ is the Cauchy loss function, and $f_{v}(g_{v}(\cdot)) = \log(1+g_{v}(\cdot)^{2}/c^{2})$, where $g_{v} = \|z^{v}-W_{v}x\|$.
It is proven in \cite{anthony2009neural} that if $f_{v}(\cdot)$ is a Lipschitz function with Lipschitz constant $L$, then the covering number of $\mathcal{F}_{v}$ is $\mathcal{N}(\mathcal{F}_{v},\epsilon) \leq \mathcal{N}(\mathcal{G}_{v},\epsilon/L)$. 
Since $L=\sqrt{2}/c$ for any view in the proposed algorithm, we have
\begin{equation}
\mathcal{N}(\mathcal{F},\epsilon) \leq \mathcal{N}(\mathcal{G},\frac{c\epsilon}{\sqrt{2}}).
\end{equation}
Considering $g_{v} = \|z^{v}-W_{v}x\|$ and  $g_{v}^{'} = \|z^{v}-W_{v}^{'}x^{'}\|$,
we have
\begin{equation}
\begin{split}
g_{v}^{'} - g_{v} = & \|z^{v}-W_{v}^{'}x^{'}\| - \|z^{v}-W_{v}x\|  \\
\leq & \|z^{v}-W_{v}^{'}x\| - \|z^{v}-W_{v}x\| \\
\leq &  \|x\|\|W_{v}^{'} - W_{v}\|.
\end{split}
\end{equation}
Given $\|x\|\leq\chi$, we suggest that $g_{v}(\cdot)$ is a $\chi$-Lipschitz continuous function w.r.t. $W_{v}$. Hence for all views, we have
\begin{equation}\label{eq:cnf}
\mathcal{N}(\mathcal{F},\epsilon) \leq \mathcal{N}(\mathcal{W},\frac{c\epsilon}{\sqrt{2}\chi}).
\end{equation}
$\mathcal{W}$ is composed of the function classes $\{\mathcal{W}_{1}, \cdots, \mathcal{W}_{m}\}$ on different views. For any combination $\{W_{1},\cdots, W_{m}\}$, the proposed multi-view learning algorithm tends to assign them with a shared coefficient $x$ for the multi-view example $\{z_{v}\}_{v=1}^{m}$, that is $[z_{1};\cdots; z_{m}] = [W_{1};\cdots; W_{m}]x$. Hence, by defining $\widetilde{W} = [W_{1};\cdots; W_{m}]$ and $\widetilde{z} = [z_{1};\cdots; z_{m}]$,
the complexity of  $\mathcal{W}$ can be approximated by that of the function class $\widetilde{\mathcal{W}}$. It is necessary to emphasize  that though we concatenate the transformation matrices to estimate the complexity of the hypotheses, it does not imply  that concatenating multiple views is  equivalent to the proposed multi-view learning algorithm. For the concatenated features $\widetilde{z}$, directly searching for $\widetilde{W}$ in a large space will increase the risk of over-fitting. By contrast,  by grouping features into different views, we are able to effectively discover the optimal matrix $W_{v}$ for each view in a smaller space with the help of the prior information on the other views.

Suppose that $\widetilde{W}$ lies in a sphere with radius $\Upsilon$. It is well known that the covering number of the sphere is upper bounded by 
\begin{equation}\label{eq:cn}
\mathcal{N}(\widetilde{\mathcal{W}}, \epsilon)\leq (\frac{3\Upsilon}{\epsilon})^{s},
\end{equation}
where the exponent $s$ is the dimension of the constraint set $\widetilde{W}$ in the sense of  its manifold structure. Finally, we complete the proof by combining Eqs. (\ref{eq:cnf}) and  (\ref{eq:cn}).

\vspace {-0.2in}
\section{Acknowledgment}
We authors greatly thank the handling Associate Editor and all the three anonymous reviewers for their constructive comments on this submission.
\ifCLASSOPTIONcaptionsoff
  \newpage
\fi
\small
\bibliography{refs}
\bibliographystyle{IEEEtran}

\vspace {-0.6in}
\tiny
\begin{biography}[{\includegraphics[width=1in,height=1.25in,clip,keepaspectratio]{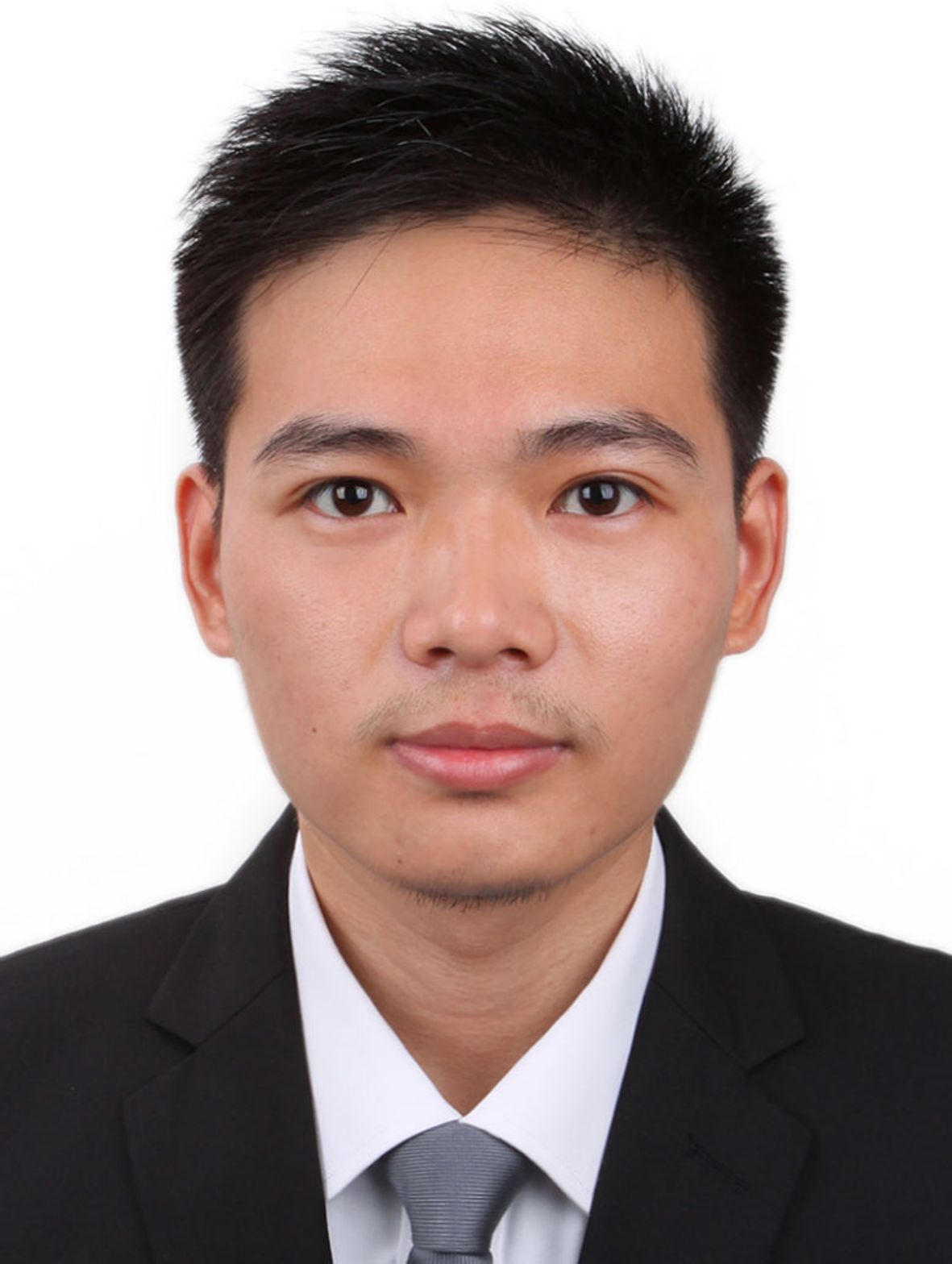}}]{Chang Xu}
 received the B.E degree from Tianjin University in 2011. Currently, he is a Ph.D candidate with the Key Laboratory of Machine Perception (Ministry of Education) in the Peking University. Previously, he was a research intern with the Knowledge Mining group in the Microsoft Research Asia. He Won the Best Student Paper Award in ACM International Conference on Internet Multimedia Computing and Service 2013. His research interests lie primarily in machine learning, multimedia search and computer vision.
\end{biography}
\vskip -0.6in
\begin{biography}[{\includegraphics[width=1in,height=1.25in,clip,keepaspectratio]{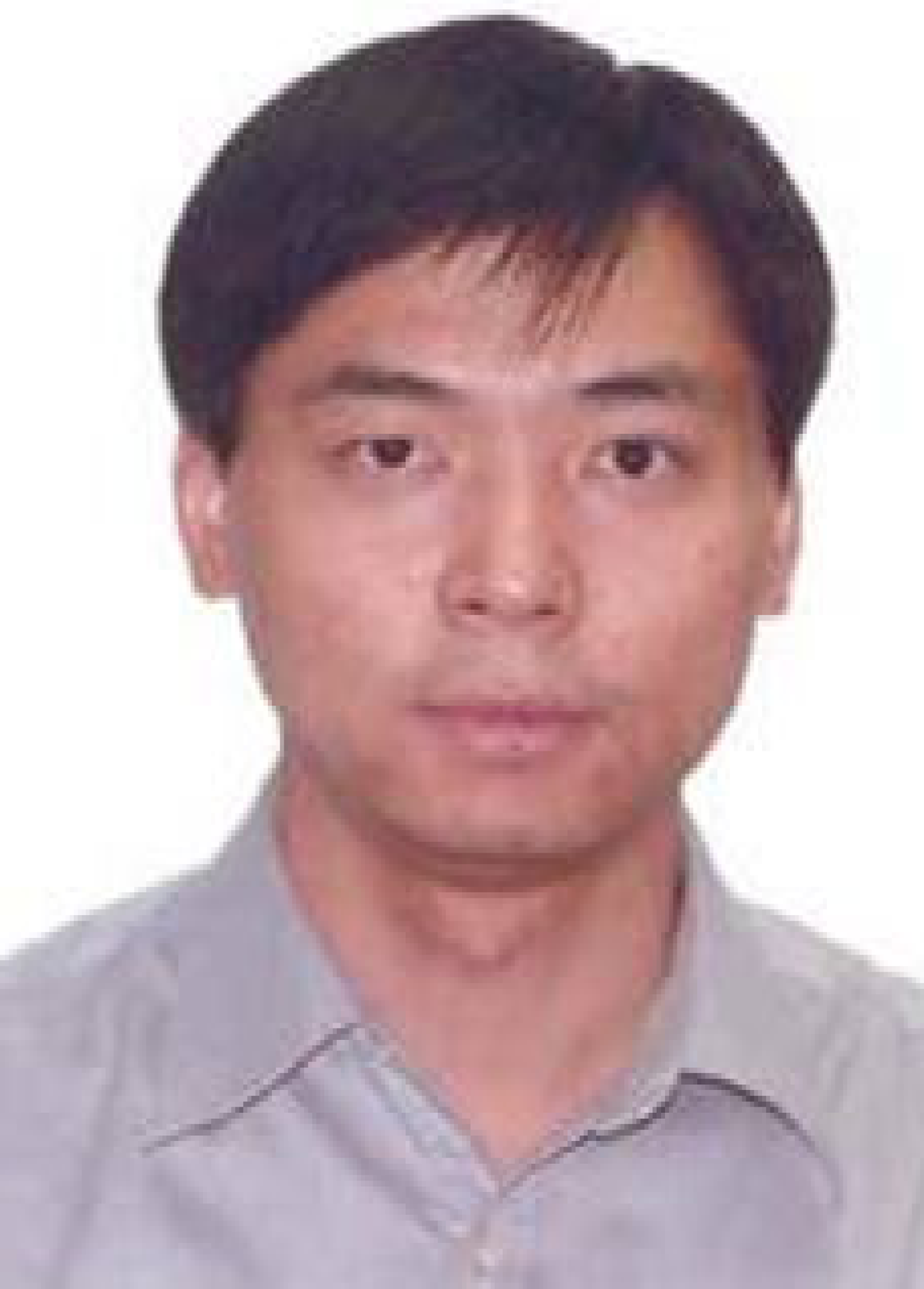}}]{Dacheng Tao} (M'07-SM'12-F'15)
 is Professor of Computer Science with the Centre for Quantum Computation \& Intelligent Systems and the Faculty of Engineering \& Information Technology in the University of Technology, Sydney. He mainly applies statistics and mathematics for data analysis problems in data mining, computer vision, machine learning, multimedia, and video surveillance. He has authored and co-authored 100+ scientific articles at top venues including IEEE T-PAMI, T-NNLS, T-IP, NIPS, ICML, AISTATS, ICDM, CVPR, ICCV, ECCV; ACM T-KDD, KDD and Multimedia,  with several best paper awards, such as the Best Theory/Algorithm Paper Runner Up Award in the IEEE ICDM’07, the Best Student Paper Award in the IEEE ICDM’13 and 2014 IEEE ICDM 10-Year Highest-Impact Paper Award. He is a Fellow of the IAPR, the IEEE, and the IET/IEE.
\end{biography}
\vskip -0.6in
\begin{biography}[{\includegraphics[width=1in,height=1.25in,clip,keepaspectratio]{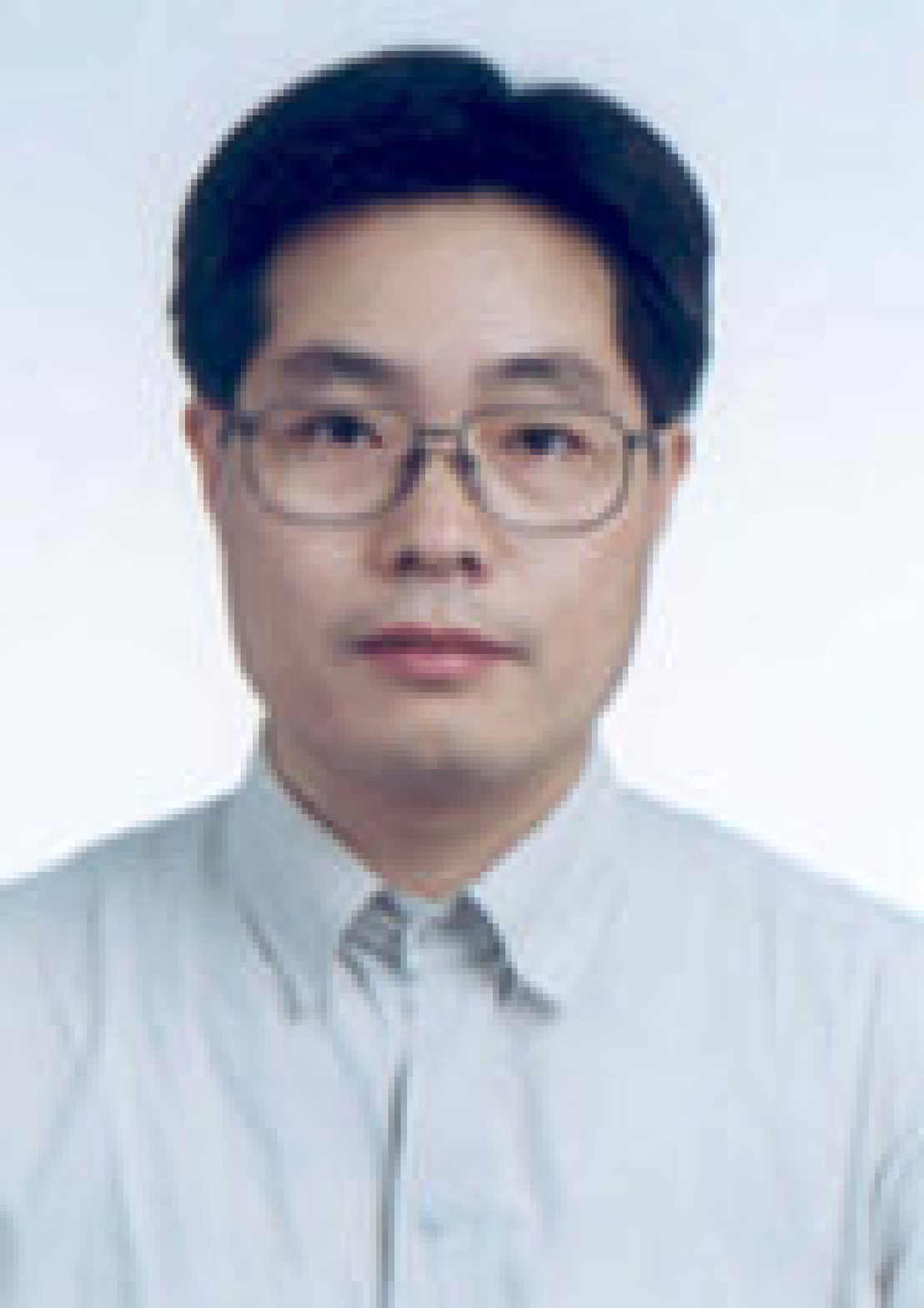}}]{Chao Xu}
received the B.E. degree from Tsinghua University in 1988, the M.S. degree from University of Science and Technology of China in 1991 and the
Ph.D degree from Institute of Electronics, Chinese Academy of Sciences in 1997. Between 1991 and 1994 he was employed as an assistant professor
by University of Science and Technology of China. Since 1997 Dr. Xu has been with School of EECS at Peking University where he is currently a Professor. His research interests are in image and video coding, processing and understanding. He has authored or co-authored more than 80 publications and 5 patents in these fields.
\end{biography}

\end{document}